\documentclass[10pt,twocolumn,letterpaper]{article}

\usepackage[pagenumbers]{cvpr} 

\usepackage{threeparttable}
\usepackage{multirow}
\usepackage{subcaption}
\usepackage{graphicx}
\usepackage{tabularx}
\usepackage{lipsum}

\newcommand{\pose}{g}
\newcommand{\K}{K}
\newcommand{\tpose}{\pose_{T}}
\newcommand{\tK}{\K_{T}}
\newcommand{\ctxt}{\ensuremath{\mathfrak{C}}}
\newcommand{\compute}{\ensuremath{\chi}}
\newcommand{\computeMLP}{\compute_{\mathtt{MLP}}}
\newcommand{\computeAttn}{\compute_{\mathtt{Attn}}}
\newcommand{\computeProj}{\compute_{\mathtt{Proj}}}

\renewcommand{\eqref}[1]{Eq.~\ref{#1}}
\newcommand{\figref}[1]{Fig.~\ref{#1}}
\newcommand{\secref}[1]{Sec.~\ref{#1}}
\newcommand{\tabref}[1]{Tab.~\ref{#1}}
\newcommand{\suppref}[1]{Supp~\ref{#1}}

\definecolor{cvprblue}{rgb}{0.21,0.49,0.74}
\usepackage[pagebackref,breaklinks,colorlinks,allcolors=cvprblue]{hyperref}

\def\paperID{REDACTED} 
\def\confName{CVPR}
\def\confYear{2026}

\title{Scaling View Synthesis Transformers}

\author{
Evan Kim\thanks{Indicates equal contribution.}\\
MIT\\
{\tt\small evnkim@mit.edu}
\and
Hyunwoo Ryu$^*$\\
MIT\\
{\tt\small hwryu@mit.edu}
\and
Thomas W. Mitchel\\
Adobe\\
{\tt\small thomas.w.mitchel@gmail.com }
\and
Vincent Sitzmann\\
MIT\\
{\tt\small sitzmann@mit.edu}
}

\begin{document}
\maketitle
\begin{abstract}

Geometry-free view synthesis transformers have recently achieved state-of-the-art performance in Novel View Synthesis (NVS), outperforming traditional approaches that rely on explicit geometry modeling. Yet the factors governing their scaling with compute remain unclear. 
We present a systematic study of scaling laws for view synthesis transformers and derive design principles for training compute-optimal NVS models. Contrary to prior findings, we show that encoder–decoder architectures can be compute-optimal; we trace earlier negative results to suboptimal architectural choices and comparisons across unequal training compute budgets. Across several compute levels, we demonstrate that our encoder–decoder architecture, which we call the Scalable View Synthesis Model (SVSM), scales as effectively as decoder-only models, achieves a superior performance–compute Pareto frontier, and surpasses the previous state-of-the-art on real-world NVS benchmarks with substantially reduced training compute. 
\href{https://www.evn.kim/research/svsm}{https://www.evn.kim/research/svsm}

\end{abstract}
\vspace{-1\baselineskip}    
\section{Introduction}
\label{sec:intro}
\begin{figure}
    \centering
    \includegraphics[width=0.9\linewidth]{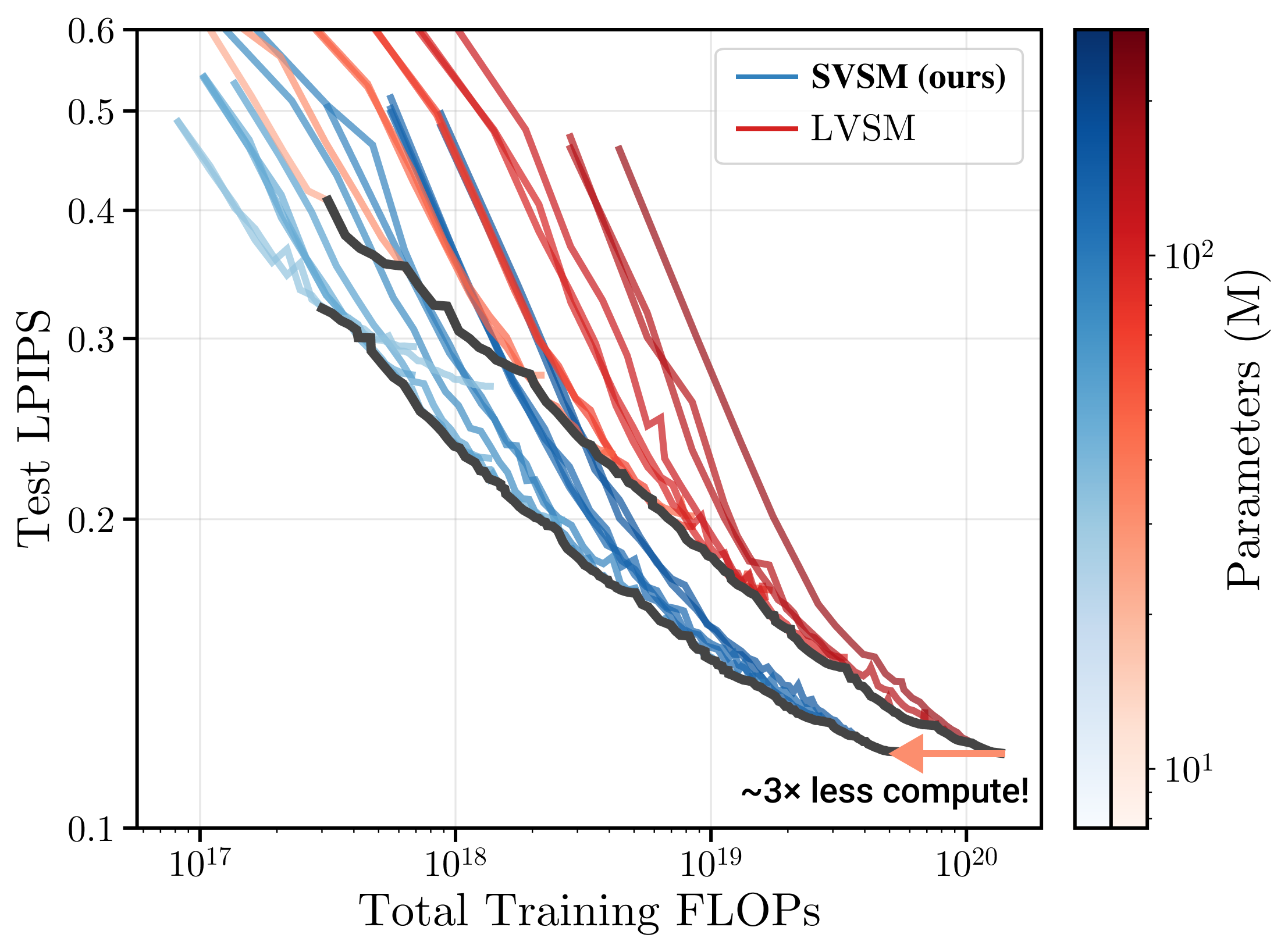}
    \caption{
    \textbf{Scaling Laws for View Synthesis Transformers.} Evaluated on RealEstate10K~\citep{zhou2018stereo}, our SVSM exhibits a $3\times$ more compute-optimal Pareto frontier than LVSM while retaining the same scaling behavior (similar slope and curvature everywhere). 
    }
    \label{fig:two-view-scale}
\end{figure}

Given a set of images of a scene with known camera poses, the goal of Novel View Synthesis (NVS) is to render novel views of the scene from arbitrary viewpoints. Single scene approaches such as NeRF~\citep{mildenhall2021nerf} and Gaussian Splatting~\citep{kerbl20233d} have achieved impressive fidelity by explicitly modeling 3D geometry and rendering. 
Feed-forward extensions of these frameworks train neural networks to reconstruct the 3D representation, achieving promising results~\citep{yu2020pixelnerf,charatan2024pixelsplat,chen2024mvsplat,zhang2024gs}.  However, their formulation inherits handcrafted 3D structure, constraining their ability to scale and handle more complex artifacts such as reflections or transparency.


Typified by Large View Synthesis Model (LVSM)~\cite{jin2024lvsm}, a new class of view synthesis models have emerged which achieve state-of-the-art rendering quality using pure transformer architectures with fewer (if any) geometric inductive biases~\cite{jin2024lvsm, jiang2025rayzer, mitchel2025true, sajjadi2022scene}. However, this class of models is relatively new, and their design space remains unexplored. In particular, training a NVS model involves making design choices over a large number of variables, including the number of context and target views, camera pose parameterizations, attention mechanisms, etc., and there does not yet exist a rigorous investigation of how these choices affect the performance, training efficiency, and inference throughput. To this point, while there exist extensive scaling analyses in language modeling and 2D vision~\cite{esser2024scaling, hoffmann2022training, kaplan2020scaling, peebles2023scalable, zhai2022scaling}, there exists no analogue for 3D vision. Thus, the goal of this work is to provide such an analysis in addition to a compute-optimal training recipe for view synthesis transformers in terms of both architecture and training strategy.

In particular, we first challenge the necessity of the decoder-only architecture proposed in LVSM~\citep{jin2024lvsm}. While powerful, it requires passing all context images through the entire transformer each time a \textit{single} target image is decoded. It is a \emph{bidirectional} model where both the target view tokens \emph{and} context view tokens are updated in each layer of the network. While this allows the model to consider only information in the context images that is relevant to the target view, it incurs substantial computational cost due to repeated processing of context views.

Instead, we advocate for an encoder-decoder design which produces an intermediate scene latent representation. This approach is potentially far more efficient: the computational cost of constructing the scene representation is amortized through repeated calls to the decoder, which efficiently extracts information from the representation via \emph{unidirectional} cross attention from scene to target.  However, the scene representation also represents an information bottleneck. Without a proper training strategy that maximally leverages the efficiency of the encoder-decoder design, it can be challenging for encoder-decoder models to outperform decoder-only models. Here, we identify that the key for unlocking the potential of encoder-decoder models lies in the way target views (\emph{i.e.} the reconstruction targets) are utilized during training. The implicit standard practice employed by prior work~\citep{charatan2024pixelsplat, jin2024lvsm, sajjadi2022scene, zhang2024gs} has been to reconstruct multiple different target views from a single scene during training. However, the consequences of this approach have never been fully analyzed. To this point, we propose and empirically validate the \emph{effective batch hypothesis}, which argues that reconstructing multiple target views per scene effectively multiplies the batch size.

These insights yield a principled transformer view synthesis model, which we call the \textit{Scalable View Synthesis Model} (SVSM), that fully capitalizes on the rendering efficiency of a unidirectional encoder-decoder architecture, maximizing training throughput without compromising performance or scalability.
We demonstrate that our unidirectional model scales as efficiently as bidirectional models, which aligns with the scalability of causal, unidirectional attention in large language models~\citep{kaplan2020scaling}. 
As part of this analysis, we also reveal scaling relationships within view synthesis that parallel those observed in the Chinchilla language model family~\citep{hoffmann2022training}.
Finally, we demonstrate that SVSM achieves state-of-the-art results in real-world NVS tasks with significantly reduced compute, challenging the previous understanding that bidirectional attention is critical to high-fidelity view synthesis~\citep{jin2024lvsm}.

\vspace{-0.5\baselineskip}
\paragraph{Key Contributions:}
\begin{itemize}
    \item We provide the first rigorous scaling analysis for novel view synthesis transformers.
    \item We propose and empirically confirm the \textit{effective batch size hypothesis} that unlocks compute-optimal training.
    \item We show that bidirectional decoding is not critical for scalable view synthesis, contrasting recent work~\cite{jin2024lvsm}.
    \item Based on this analysis, we present a compute-optimal model that achieves a new state-of-the-art in real-world NVS tasks with substantially reduced training compute. 
\end{itemize}




\section{Related Work and Preliminaries}
\label{sec:background-scale}

%
\paragraph{Generalizable novel view synthesis.} 
In generalizable novel view synthesis, we are given a set of $V_C$ context images paired with known camera poses and intrinsics $\ctxt = \left\{(I_i, \, \pose_i, \, \K_i) \mid i = 1, \, \ldots, \,V_C\right\}$. The typical objective is to synthesize an \emph{unseen} view of the same scene given a target camera configuration $\tpose, \tK$:
\begin{equation}\label{eq:view_synthesis}
\tilde{I}_T = \mathrm{Render}\big[\ctxt,  \tpose,  \tK \big].
\end{equation}

\begin{figure*}[t]
    \centering
    \vspace{-0.8cm}
    \includegraphics[width=0.9\textwidth]{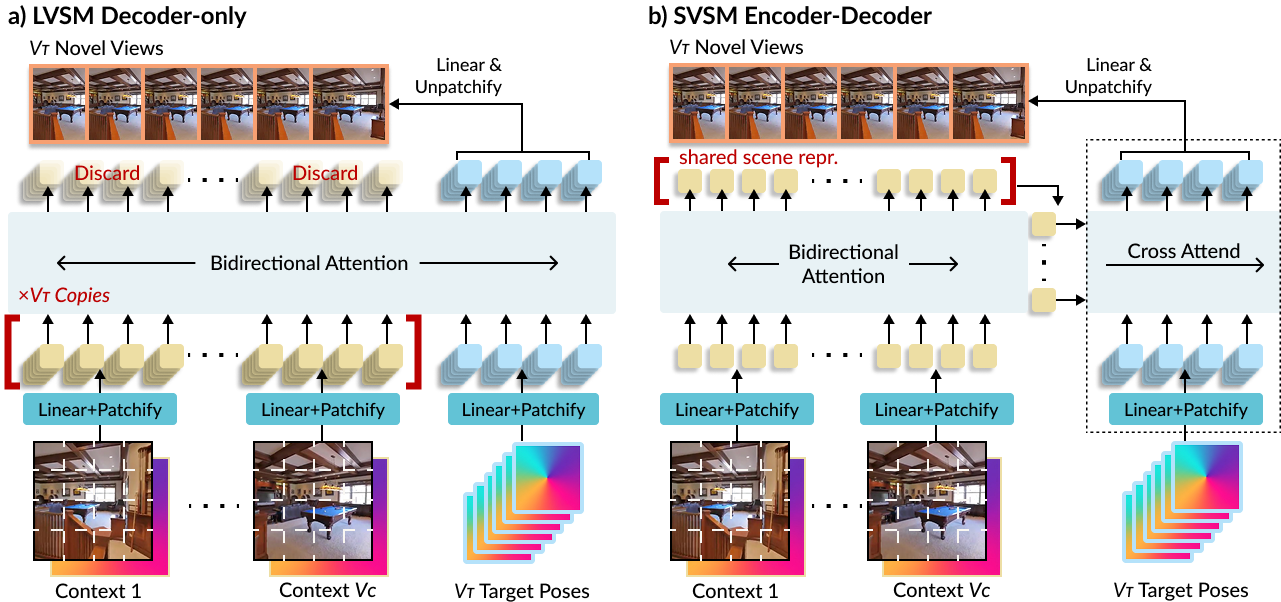}
    \caption{
        \textbf{Architectures of the current SOTA, the decoder-only LVSM~\citep{jin2024lvsm} (a) and SVSM (ours, b).} Our cross-attention based decoder enables parallel rendering of multiple target views after a single scene encoding. Each target view is decoded independently given the shared scene representation, but the cross-attention allows these independent decodings to be executed in parallel.
        \vspace{-1\baselineskip}
    }
    \label{fig:lvsm-svsm}
\end{figure*}

One line of work attempts to solve this problem with neural network architectures that explicitly model aspects of 3D image formation, for instance, via differentiable rendering or using epipolar line constraints~\cite{yu2020pixelnerf,sitzmann2019srns,charatan2024pixelsplat,szymanowicz2024splatter,suhail2022generalizable}. 
In contrast, \emph{geometry-free} methods avoid explicit geometric modeling in favor of flexibility and generality~\cite{eslami2018neural,sitzmann2021lfns,sajjadi2022scene,rombach2021geometry, sajjadi2023rust}. Here, we are primarily interested in a recently proposed subclass of these models which achieve state-of-the-art results with pure transformer architectures~\cite{jin2024lvsm, jiang2025rayzer, sajjadi2022scene, mitchel2025true}. In particular, we seek to study the ``Large View Synthesis Model'' (LVSM)~\cite{jin2024lvsm}, which achieves state-of-the-art NVS performance and serves as the prototypical instance of the view synthesis transformer. 
LVSM can be implemented in two ways: as either an encoder-decoder model or decoder-only model. The authors' proposed decoder-only variant is far more performant, so we primarily consider this architecture (pictured in \figref{fig:lvsm-svsm}, left) in our analysis.

Decoder-only LVSM consists of a single module: A decoder $\mathcal D $ which ingests the raw context $\mathfrak C$ along with a \emph{single} target configuration, and aims to render a prediction of the target view $\tilde{I}_T =  \mathcal D \big[\mathfrak C, \tpose, \tK\big]$. This model is \emph{bidirectional} as the context view tokens are updated with information about target pose and tokens in each layer. Thus, the processed context tokens cannot be reused and must be re-initialized and updated each time a new target view is rendered. As a consequence, rendering $V_T$ target views requires $V_T$ forward passes through the full network. 
%
Decoder-only LVSM consists of standard ViT layers~\citep{dosovitskiy2020image} which apply self-attention ($\mathtt{Attn}$) followed by an MLP ($\mathtt{MLP}$). 
Therefore, the FLOPs on a forward pass scale linearly with the number of target views $V_T$
\begin{equation}\label{eqn:lvsm-flops}
\begin{split}
    \computeMLP^{\text{(LVSM)}} & \propto V_T\times (V_C+1)
    \\
    \computeAttn^{\text{(LVSM)}} & \propto V_T\times (V_C+1)^2
\end{split}
\end{equation}
where $\computeMLP^{\text{(LVSM)}}$ and $\computeAttn^{\text{(LVSM)}}$ are the FLOPs consumed by the MLP and attention in the decoder-only LVSM. In our studies, $\computeMLP^{\text{(LVSM)}}$ is the dominating factor (for more details, see \suppref{supp:flop-calcs}). In what follows, we will continue to use $\chi$ to denote the compute metric, typically measured in FLOPs.

\newpage

\paragraph{Scaling Laws.}
As the scale of deep learning models continues to increase, it has become increasingly important to understand the relationship between performance and compute to ensure an efficient use of resources. To this end, scaling analyses have been conducted for language models~\citep{hoffmann2022training, kaplan2020scaling}, vision transformers~\citep{zhai2022scaling}, and diffusion transformers~\citep{esser2024scaling, peebles2023scalable}. The general approach is straightforward:  train models at different compute budgets and analyze performance as a function of compute. 

Scaling studies are useful in two ways. First, they provide a predictable trend of performance with compute, essentially describing a performance metric $P$ as function of compute \compute{} which can reveal characteristic scaling behavior. For example, in language models, $P(\compute{})$ has been found to approximately follow a power-law~\citep{kaplan2020scaling}. Second, they have revealed which hyperparameter choices are most effective as models scale. For instance,  Chinchilla scaling laws~\cite{hoffmann2022training} reveal the best way to trade-off between model size $N$, measured in parameter count, and the number of training samples used, $D$. Analysis is performed by sweeping across a wide range of $N$ and $D$ to discover for compute budget $\compute{}$ the optimal $N$ and $D$. Then, taking this paired data and assuming a power law relation between $N_{\text{opt}}$ and $D_{\text{opt}}$ and $\compute{}$, one can fit powers $a$ and $b$ of $\compute{}$ to the pairings
\begin{equation}\label{eqn:chinchilla}
    N_{\text{opt}}(\compute{}) \propto \compute{}^a, D_{\text{opt}}(\compute) \propto \compute{}^b.
\end{equation}
Remarkably, experiments demonstrate $a \approx b$, suggesting  $N$ and $D$ should scale proportionally. In our study, we will focus on the second of these two kinds of studies: replicating the Chinchilla study (\secref{sec:scaling-two-view}) in the NVS domain and exploring other similar tradeoffs, such as the effective batch size (\secref{sec:effective-batch})

\paragraph{Extremely long context view synthesis.}
Recently, there has been a line of work aiming to develop view-synthesis and 3D-reconstruction models whose computational cost scales linearly with the number of context images~\citep{zhang2025test,wang2025continuous, ziwen2025llrm}. Indeed, as in \eqref{eqn:lvsm-flops}, as $V_C$ grows, the quadratic cost of attention comes to dominate the compute and becomes infeasible. In that regime, such linear-cost models are promising alternatives. However, we restrict our focus to sparse to moderately sparse view synthesis, for which linear-cost models are currently not state of the art.





\definecolor{B128first}{HTML}{F6C928}
\definecolor{B256}{HTML}{36827F}
\definecolor{B128second}{HTML}{96A654}
\definecolor{B1024}{HTML}{464D77}

\begin{figure*}[t]
    \centering
    %
    \includegraphics[width=\textwidth]{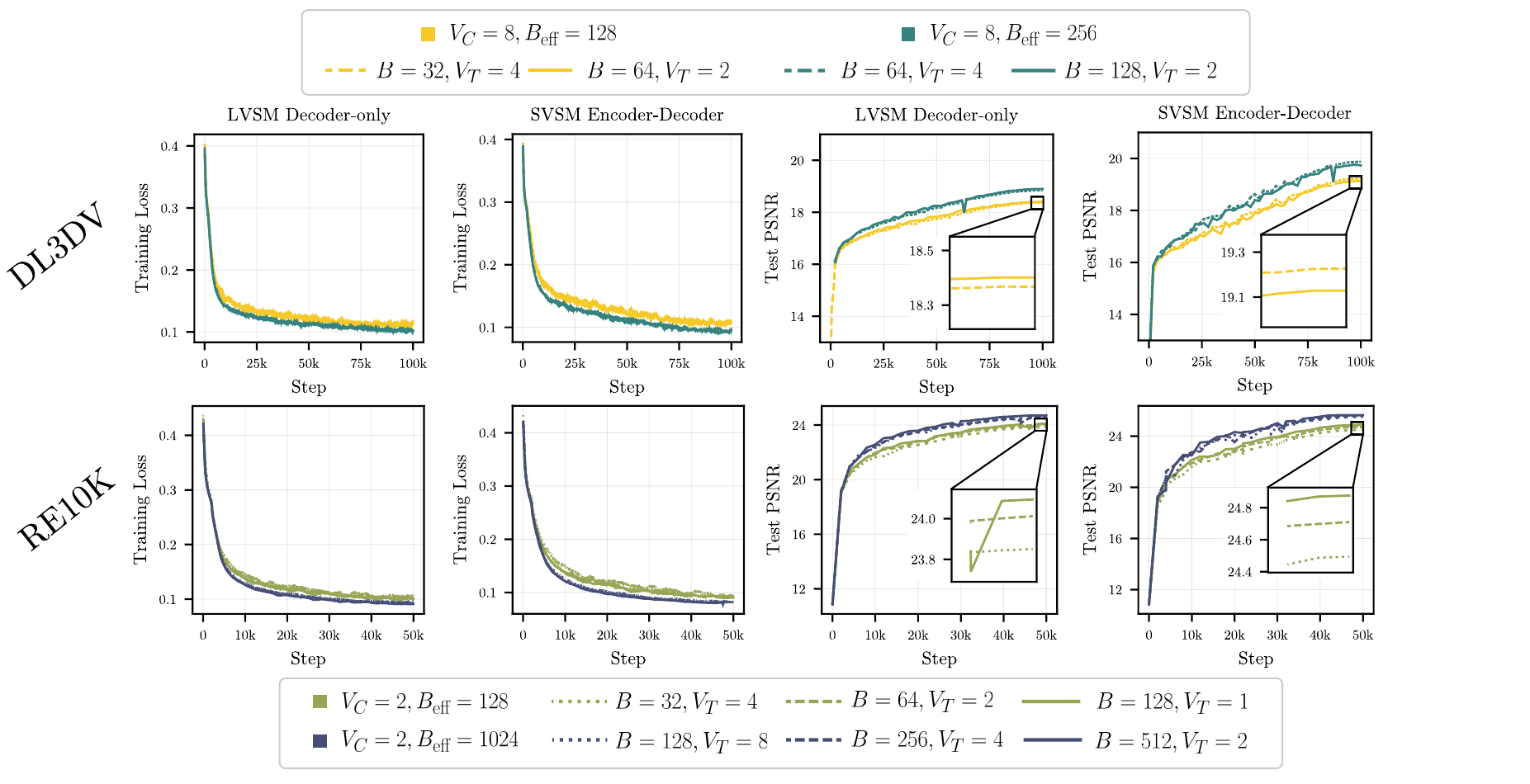}
    \vspace{-0.8cm}
    \caption{\textbf{Effective Batch Size}. Training loss (smoothed with a rolling-average) and test PSNR measured throughout training across various paired $B$ and $V_T$ runs provide evidence for our effective batch size hypothesis: Models trained with the same product of number of scenes in the batch $B$ and number of reconstruction target views $V_T$, \emph{i.e.} runs with the same \emph{effective batch size} $B_{\text{eff}}$, perform the same and are colored identically. On $V_C = 8$ \textbf{(top)}, we sweep across $B_{\text{eff}} = $ \textcolor{B128first}{128}, \textcolor{B256}{256} on DL3DV, and on $V_C = 2$ \textbf{(bottom)}, we sweep across $B_{\text{eff}} =$ \textcolor{B128second}{128}, \textcolor{B1024}{1024} on RealEstate10K.
    \vspace{-1\baselineskip}
} 
    \label{fig:effective-batch} 
\end{figure*}
\section{Encoder-Decoder View Synthesis}
\label{sec:architecture}

As discussed in \secref{sec:intro} and \secref{sec:background-scale}, the decoder-only LVSM may not be the most compute-optimal due to the recomputation per-target view rendered. 
This motivates us to seek an alternative model with a fixed scene representation that can be decoded in a purely \emph{unidirectional} manner (\emph{i.e.} via cross attention) to reduce the cost of rendering and avoid redundant reprocessing of context information. To this extent, we introduce the Scalable View Synthesis Model (SVSM), which can be viewed as a simple modification to encoder-decoder LVSM.
Specifically, our architecture implements \eqref{eq:view_synthesis} by first processing the context set $\mathfrak{C}$ with a transformer encoder, producing a set of latent tokens (a ``scene representation'') $\mathbf{z} = \mathcal E[\mathfrak{C}]$.
The encoder $\mathcal E$ is standard transformer with full bidirectional self-attention.
Unlike encoder-decoder LVSM, we do not employ a fixed-size scene representation, but instead take the set of encoded context image patch tokens as the scene representation to avoid introducing a bottleneck. To render a novel view, a cross-attention based decoder $\mathcal D$ ingests the target configuration and the fixed scene latent tokens $\mathbf{z}$ to render the target view, $\tilde{I} = D \big[\mathbf{z}, \, \tpose, \tK \big]$. 
To render multiple images of the same scene, we only require encoding the context set once, re-using the scene embedding $\mathbf{z}$. As with LVSM, each novel view is decoded independently given $\mathbf{z}$ (i.e., there is no interaction between target views). However, because the decoder uses cross-attention rather than bidirectional self-attention over all tokens, these independent target views can be decoded in parallel, without redundant recomputation of the scene representation.
To be more concrete, this architecture reduces the complexity of rendering $V_T$ targets to
\begin{equation}\label{eqn:svsm-flops}
\begin{split}
    \computeMLP^{\text{(SVSM)}} & \propto V_T + V_C
    \\
    \computeAttn^{\text{(SVSM)}} & \propto V_C\times (V_T + V_C).
\end{split}
\end{equation}
In other words, assuming most of the compute is due to the MLP layers (see \suppref{supp:flop-calcs}), rendering $V_T$ target views requires $\mathcal O(V_T + V_C)$ FLOPs. In the limit of inference where $V_T \gg V_C$, this reduces to $\mathcal O(V_T)$, in stark contrast to the $\mathcal O(V_TV_C + V_T)$ of LVSM (see \eqref{eqn:lvsm-flops}). Further, the benefit of this paradigm extends beyond inference. As long as we are training with multiple target views, as is standard practice~\citep{charatan2024pixelsplat, jin2024lvsm, sajjadi2022scene, zhang2024gs}, the parallel nature of unidirectional decoding can save substantial training compute.


However, this reduction comes with a cost. Unlike LVSM, our encoder cannot proactively discard information irrelevant to the target view but instead needs to encode all necessary information for rendering \emph{any} target view. Indeed, parameter count and training steps being equal, SVSM performs worse than LVSM. However, as we will show, SVMS’s amortized rendering enables us to dramatically increase its size and training steps, such that when normalized by compute budget, SVSM significantly outperforms LVSM.



\section{The Effective Batch Size for View Synthesis}
\label{sec:effective-batch}

As the cost of a forward pass scales both with the number of different scenes (the batch size $B$) as well as the number of target views ($V_T$) that we seek to render per scene, this introduces an additional hyperparameter into the training regime: What is the \emph{optimal} trade-off between the number of target views and the number of different scenes? 
We study this question empirically and reveal that what matters is the \emph{product} of target views and batch size, which we call the \emph{effective batch size} of a NVS model. 
\vspace{-1\baselineskip}

\paragraph{Analysis Setup.}
We define effective bath size as $B_{\text{eff}} \equiv B \cdot V_T$, where $B$ is the number of scenes in a training batch, and $V_T$ is the number of rendering targets used per training scene.
We train both decoder-only LVSM and the proposed SVSM models across two datasets --- DL3DV~\cite{ling2024dl3dv} and RealEstate10K~\cite{zhou2018stereo} (RE10K) --- with $V_C = 8$ and $V_C = 2$ while holding $B_{\text{eff}}$ constant and varying $B$ and $V_T$. Specifically, we test $B_{\text{eff}} = 128$ on both datasets, and additionally test $B_{\text{eff}} = 1024$ on RE10K and $B_{\text{eff}} = 256$ on DL3DV. For DL3DV we use the official test-train split, and for RE10K, we use the pixelSplat~\citep{charatan2024pixelsplat} test-train split.
Further training details are outlined in \suppref{supp:exp-details}.


\paragraph{Effective Batch Size is What Matters.} Results for all training runs are shown in \figref{fig:effective-batch}. Remarkably, in all cases -- across both models, both $V_C$ settings, and all $B_{\text{eff}}$ sets -- the test metric and the training loss behavior remain approximately constant along a $B_{\text{eff}}$-level set. This effect is especially clear in the $V_C = 8$ case, where the test PSNR varies by at most $\pm 0.1$ and remains present in the $V_C = 2$ case, where the variation is at most $\pm 0.2$~PSNR. Tuning $B$ and $V_T$ within the same effective batch size $B_{\text{eff}}$ does not result in significant difference in the final performance outcome, we exclude this degree of freedom from our subsequent analyses and treat $B_{\text{eff}}$ as the true batch size. 




\paragraph{SVSM Enables Compute-Optimal Tradeoff.}
How can we interpret this result through the lens of compute-optimiality? For the LVSM decoder-only model, the training compute scales as 
\begin{equation}
    \compute{}^{\text{(LVSM)}} \propto B V_T (V_C + 1) = B_{\text{eff}}(V_C + 1).
\end{equation}
Thus, any training settings within constant $B_{\text{eff}}$ not only achieve within-noise results (as per our effective batch result), but also require the same number of FLOPs. This means for the decoder-only model there is \emph{no advantage to be gained by tuning $V_T$}. In contrast, for the SVSM model, training compute is proportional to
\begin{equation}
    \compute{}^{\text{(SVSM)}} \propto B(V_C + V_T) = B_{\text{eff}} + BV_C.
\end{equation}
Therefore, by reducing $B$ and increasing $V_T$, one can achieve the same effective batch size -- and consequently, the same performance -- with lower compute cost. This justifies our original motivation for a model design that efficiently decodes multiple $V_T$.


%

%

\begin{table*}[t]
  \centering
  \begin{threeparttable}
    \scriptsize
    \setlength{\tabcolsep}{3pt} 

    \begin{subtable}{\linewidth}
      \begin{tabular*}{\linewidth}{@{\extracolsep{\fill}} l *{9}{c}}
        \toprule
        
        & \multicolumn{3}{c}{\textbf{Scale Parameters}}
        & \multicolumn{3}{c}{\textbf{Reconstruction Quality}}
        & \multicolumn{3}{c}{\textbf{Rendering FPS (↑)}} \\
        
        \cmidrule(lr){2-4} \cmidrule(lr){5-7} \cmidrule(lr){8-10}
        
        \textbf{Model} 
        & \textbf{Model Size} & \textbf{Train Iters} & \textbf{Train FLOPs (↓)}  
        & \textbf{PSNR} (↑) & \textbf{SSIM} (↑) & \textbf{LPIPS} (↓)
        & $V_C$=2  & $V_C$=4  & $V_C$=8  \\
        
        \midrule
        
        LVSM Encoder-Decoder~\citep{jin2024lvsm}
        & 173M & 100k & 2.53 zflops 
        & 28.58 & 0.893 & 0.114
        & \underline{53.7} & \underline{52.9} & \textbf{52.7} \\
        
        LVSM Decoder-Only~\citep{jin2024lvsm}
        & 171M & 100k & 1.60 zflops 
        & 29.67 & 0.906 & \underline{0.098}
        & 37.9 & 19.5 & 8.6 \\
        
        SVSM Enc-Dec (\textbf{ours}, Iter-matched)
        & 740M & 100k & \textbf{0.74 zflops} 
        & \underline{29.80} & \underline{0.907} & \underline{0.098} 
        & 48.6 & 42.7 & 35.0 \\

        SVSM Enc-Dec (\textbf{ours}, Pareto-optimal)
        & 416M & 170k & \textbf{0.77 zflops} 
        & \textbf{30.01} & \textbf{0.910} & \textbf{0.096}
        & \textbf{71.0} & \textbf{61.8} & \underline{49.7} \\
        
        \bottomrule
      \end{tabular*}
      \label{tab:my_new_table_part1} 
    \end{subtable}

  \caption{\textbf{Stereo (}$V_C=2$\textbf{) NVS Results of the Largest Models.} 
  All models use a patch size of $8$ with input images at $256 \times 256$ resolution. Our models achieve the highest reconstruction metrics while using less than half of the training compute. The rendering FPS of both SVSM models is also much faster than that of the LVSM decoder-only model, though both are slower than the LVSM encoder-decoder when $V_C$ is large.}
  \label{tab:two-view-result} 
  \end{threeparttable}
  \vskip -1.0\baselineskip 
\end{table*}

\section{Scaling Laws for Stereo \boldmath\texorpdfstring{$\left(V_C{=}2\right)$}{(Vc=2)} NVS}
\label{sec:scaling-two-view}

\textbf{Analysis Setup.}
We first experiment in the most classical setting for feed-forward Novel View Synthesis -- stereo synthesis with two context views. All training and evaluation for the $V_C = 2$ case is done on RealEstate10K~\cite{zhou2018stereo}. As before, we follow the test-train split of pixelSplat~\cite{charatan2024pixelsplat}, along with the same evaluation framework. For training, we use $V_T = 6$ target views per training example, following the setup of~\citep{jin2024lvsm}. We use a batch size of $256$ for all experiments. We use a patch size of $p = 16$ for all experiments, except in the case of table \ref{tab:two_view_STOA}, where we use $p = 8$ to compare against the reported state-of-the-art numbers from LVSM. To ensure stable scaling of both models, we also apply a $1/\sqrt{L}$ multiplier to the residuals, where $L$ is the depth of the transformer, following ideas from depth-$\mu$P~\citep{yang2023tensor,bordelon2023depthwise}. Further training details can be found in the supplementary material. We use the test LPIPS~\citep{zhang2018unreasonable} loss as our primary performance metric, as it produces near linear trends on log-log plots against FLOPs. 
\vspace{-1\baselineskip}

\label{subsec:scalinglaws}

\begin{figure}
    \centering
    \includegraphics[width=\linewidth]{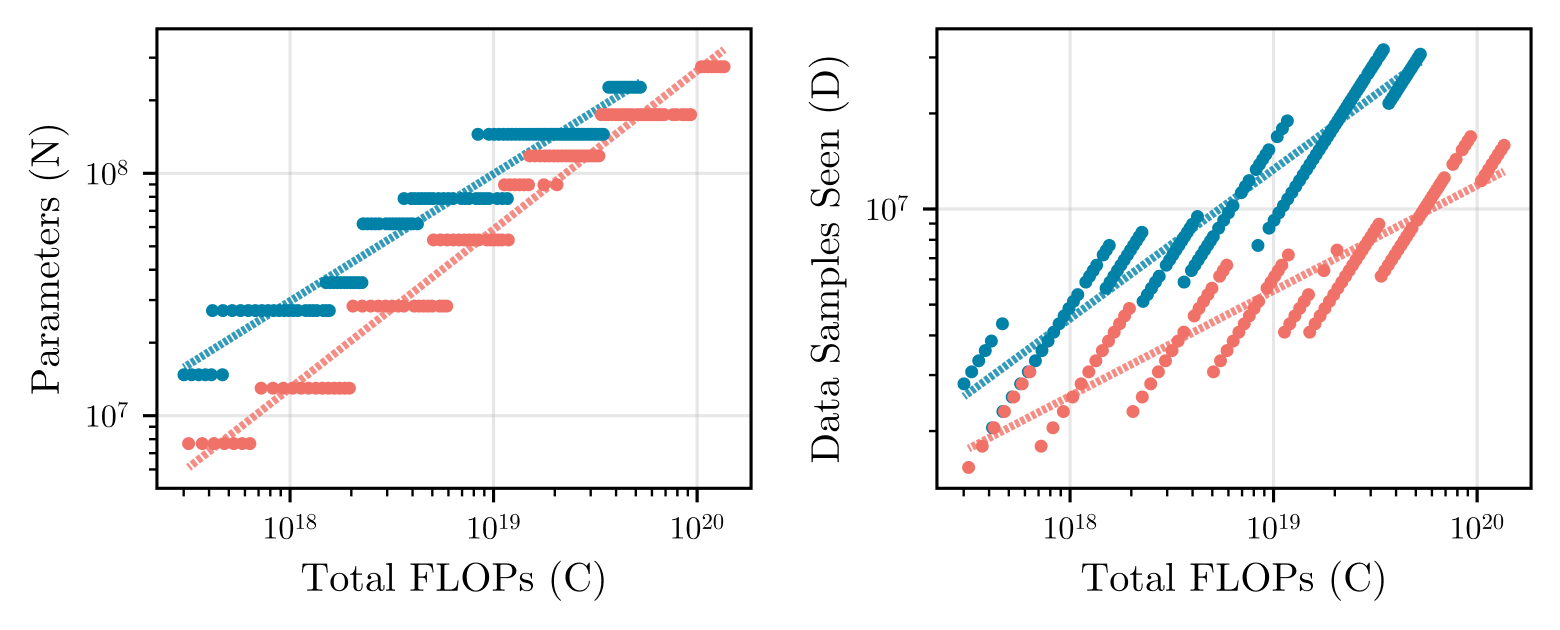}
    \vspace{-0.5cm}
    \caption{
        \textbf{Data and Model Scaling Plots.} 
        While our model~(\textcolor[HTML]{0081A7}{blue}) is optimal when sufficient data is available, decoder-only LVSM~(\textcolor[HTML]{F07167}{red}) performs better with less data. The Pareto frontier analysis shows that our model is more data-hungry. Our model is also less parameter-efficient, although the gap closes as we increase the training compute. However, with sufficient data and compute, our model~(\textcolor[HTML]{0081A7}{blue}) is overall superior in terms of training compute-optimality and rendering speed.
        \vspace{-1\baselineskip}
    }
    \label{fig:opt-model-scaling}
\end{figure}

\paragraph{Scaling Laws.}
We now follow the approach in language modeling~\cite{hoffmann2022training} to answer the question: \textit{for a given compute-budget, what is the optimal performance that can be attained?} For both model families, we sweep across a range of models from around 7M to 300M parameters, training each model for 3-4 different sample counts to densely cover the  FLOP range~\citep{hoffmann2022training}. Our training runs span a compute range of $10^3$ magnitudes: 100 petaflops to 100 exaflops. From this data, we are then able to determine a mapping from compute budgets $C$ to minimum test LPIPs -- the Pareto frontier. 

\begin{figure*}[t]
    \centering
    \includegraphics[width=\textwidth]{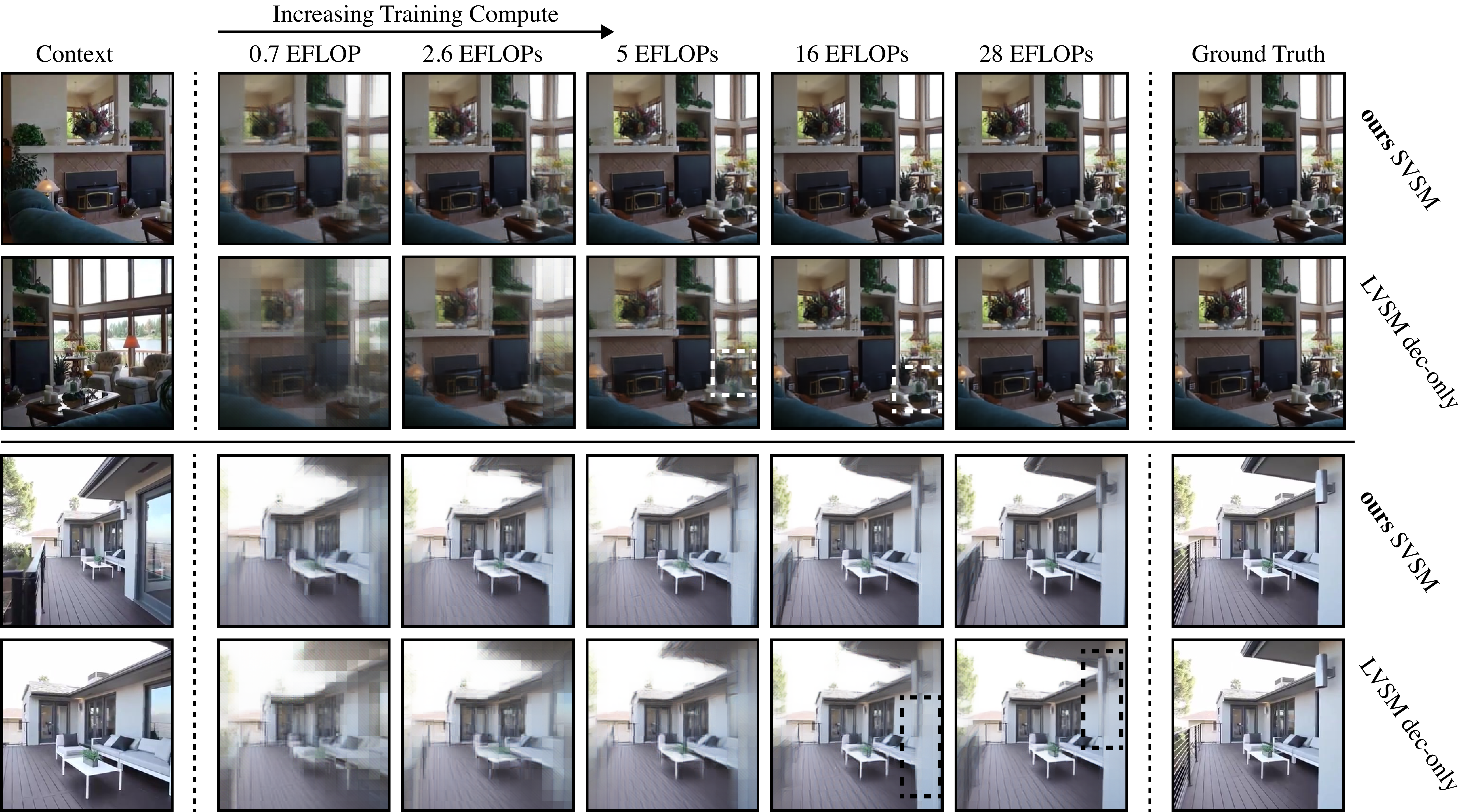}
    \caption{
        \textbf{Qualitative Scaling Behavior}, $V_C = 2$. From left to right, both models steadily increase in rendering quality until reaching near photo-realistic results. Compared vertically, for a given compute-budget, SVSM renderings consistently contain less artifacts.
        \vspace{-1\baselineskip}
    }
    \label{fig:qualitative-two-view}
\end{figure*}

We plot results in~\figref{fig:two-view-scale}, with their Pareto frontiers marked in dark gray. Plotted on a log-log scale, both models exhibit a consistent downward trend on test LPIPS with more compute. More significantly, the Pareto frontiers of both families have approximately the same slope at points of the same performance (see \suppref{supp:fits}), and SVSM's frontier is shifted left by a factor of $3$. Thus, as an initial result, our scaling laws show us that our encoder-decoder architecture \textit{scales exactly the same as the decoder-only LVSM while using $3\times$ less training-compute.} Qualitative results of this scaling are shown in \figref{fig:qualitative-two-view}, and we see that when FLOP-matched, SVSM has better rendering quality.

\vspace{-1\baselineskip}

\begin{table}[t]
  \centering
  \begin{threeparttable}
    \scriptsize
    \setlength{\tabcolsep}{3pt} 

    \begin{tabular*}{\columnwidth}{@{\extracolsep{\fill}} l ccc} 
      \toprule
      
      & \multicolumn{3}{c}{\textbf{Recon Quality}} \\
      
      \cmidrule(lr){2-4} 
      
      \textbf{Model} 
      & \textbf{PSNR} (↑) & \textbf{SSIM} (↑) & \textbf{LPIPS} (↓) \\
      
      \midrule
      
      pixelNeRF~\citep{yu2020pixelnerf} & 20.43 & 0.589 & 0.550 \\
      pixelSplat~\citep{charatan2024pixelsplat} & 26.09 & 0.863 & 0.136 \\
      MVSplat~\citep{chen2024mvsplat} & 26.39 & 0.869 & 0.128 \\
      GS-LRM~\citep{zhang2024gs} & 28.10 & 0.892 & 0.114 \\
      SVSM Enc-Dec \textbf{(ours)} & \textbf{30.01} & \textbf{0.910} & \textbf{0.096} \\
      
      \bottomrule
    \end{tabular*} 
    
    \caption{
        \textbf{Comparison to Geometry-Aware Methods.}
        Our method achieves a new state-of-the-art on RealEstate10K~\citep{zhou2018stereo} with the set from~\citet{charatan2024pixelsplat}, outperforming not only LVSM, but also prior work with explicit 3D structure. 
    }
    \label{tab:two_view_STOA}

  \end{threeparttable}
      \vspace{-1\baselineskip}

\end{table}

\paragraph{Optimal Model Choice.}

\begin{table}[h]
\footnotesize
\centering
\begin{tabular}{lcc}
\toprule
\textbf{Model} & Parameter Coeff.\ $a$ & Data Coeff.\ $b$ \\
\midrule
LVSM & 0.65 & 0.33 \\
SVSM & 0.52 & 0.47 \\
\bottomrule
\end{tabular}
\caption{\textbf{Parameter and Data Scaling Coefficients.} As regressed from the plots in \figref{fig:opt-model-scaling}, we find power law coefficients for scaling models and data with respect to compute.
\vspace{-1\baselineskip}
}
\label{tab:opt-model-table}
\end{table}

From our scaling experiments, we can further extract a compute-optimal training recipe for our view synthesis transformers, as demonstrated by~\citet{hoffmann2022training}. For each compute budget $\compute{}$, we determine the corresponding optimal model size $N$ and the amount of training data $D$ used at that point. Then, plotting $N$ and $D$ against $\compute{}$, we can extract the Chinchilla scaling equations (\eqref{eqn:chinchilla}) by fitting lines onto the log-log plots in Fig. \ref{fig:opt-model-scaling}. The recovered coefficients are shown in \tabref{tab:opt-model-table}, which inform how to train models which end on the frontier.

From these results it follows that for SVSM, if we increase our compute budget by a factor of $k\times$, it should approximately be equally allocated between increasing the model size by $\sqrt k$ and increasing data sample count by $\sqrt k$ as $a_{\text{SVSM}} \approx b_{\text{SVSM}}$, matching the findings of the Chinchilla scaling laws~\citep{hoffmann2022training} for language models. For LVSM this relationship does not seem to hold exactly, but the fit still shows that requires significant scaling of data with respect to compute, though to a smaller power. 

\begin{figure*}[t]
    \centering
    \includegraphics[width=1.0\textwidth]{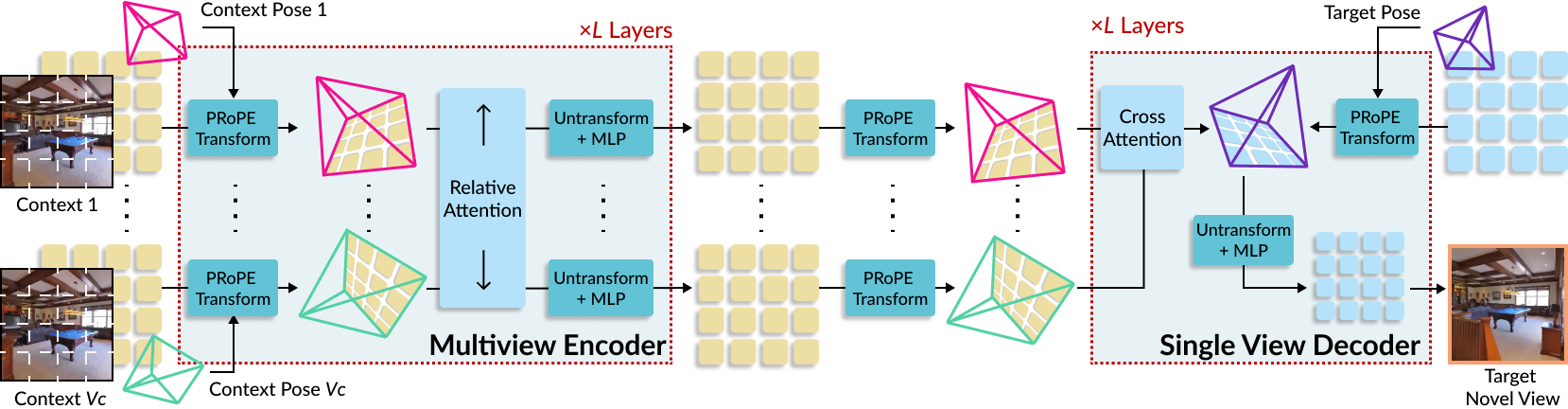}
    \caption{\textbf{Multiview PRoPE.} We find that multiview projective RoPE embeddings~\citep{miyato2023gta,kong2024eschernet,li2025cameras} are critical for our model to scale with compute and data in the multiview setting ($V_C>2$). For each layer of the multiview transformer encoder, PRoPE embeddings use context camera poses to transform context view tokens into a common coordinate frames before the attention layer, and apply the inverse transformation before each MLP. To render, both context features and query tokens of the target view are transformed by PRoPE before cross-attention.}
    \vspace{-\baselineskip}
    \label{fig:prope}
\end{figure*}
\begin{figure}
    \centering
    \includegraphics[width=\linewidth]{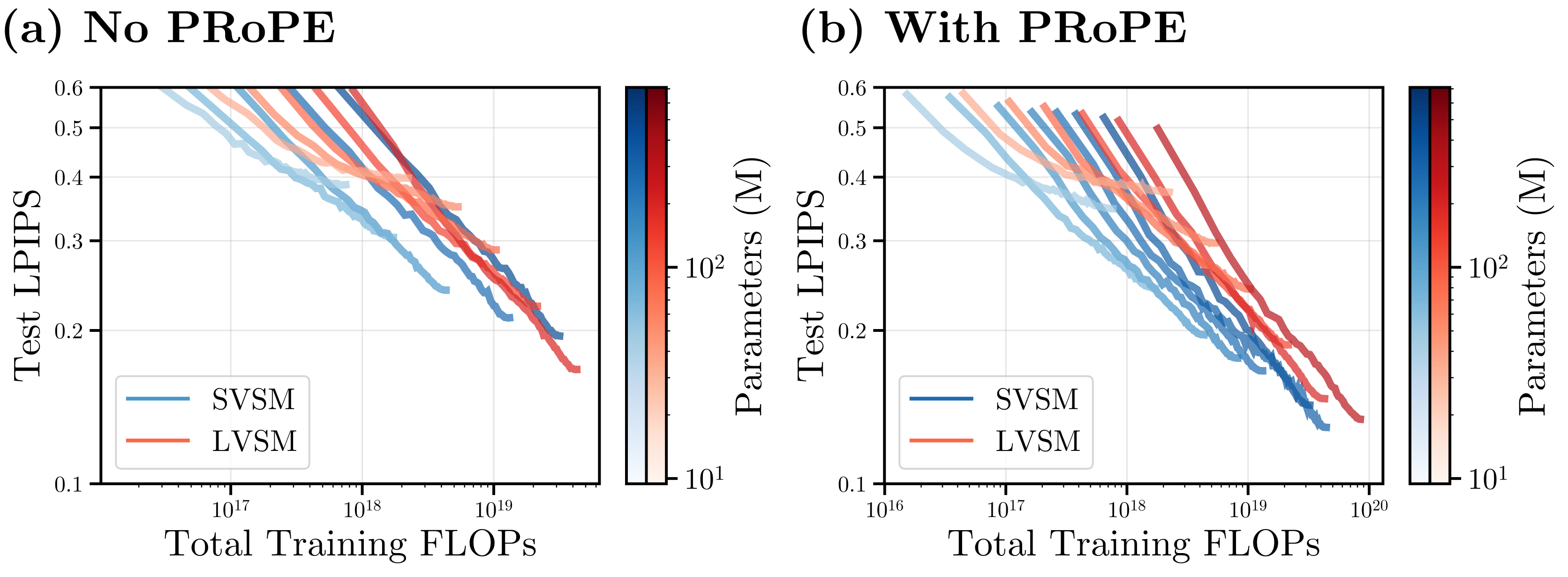}
    \caption{\textbf{Multiview Scaling Behavior.} Conducted on DL3DV~\cite{ling2024dl3dv}. \textbf{(a)} For $V_C{>}2$, without PRoPE, SVSM saturates and stops scaling much more quickly than LVSM. \textbf{(b)} When PRoPE is added, SVSM continues scaling with a better Pareto-frontier.}
    \label{fig:mv_scale}
\end{figure}

Notably our data sample counts include \emph{repeated scene data}, as we only have access to small, pose-labeled datasets. This differs from standard scaling practice, in which models are typically trained for less than one full epoch~\citep{hoffmann2022training, kaplan2020scaling, zhai2022scaling}. Although we have not yet seen evidence of overfitting in our experiments -- perhaps due to the diversity of view sequences which are sampled during training -- we have shown that increasing scale requires increasing the number data samples. Thus, having access to larger amounts of diverse posed data will be essential for developing large-scale generalizable NVS models.
\vspace{-1\baselineskip}






\paragraph{SVSM-420M/740M Results.}
\looseness=-1
Finally, combining our scaling law findings, we train two separate models --- SVSM-420M and SVSM-740M, aptly named to denote their parameter counts --- to compare against the original results of LVSM's largest model on RealEstate10K. Due to compute-constraints, we train our models at a lower total budget of around $10^{21}$ FLOPs and a batch size of $256$,  approximately half the FLOPs and exactly half the batch size used by LVSM.  We train two models under this budget: (1) a flop-matched model with the 24 layer LVSM model for a forward pass of a single training sample with $V_C = 2, V_T=  6$ and; (2) A model whose parameter count is given by plugging the budget\footnote{To be more specific, we plug $\compute/4$ into the scaling law, in order to adjust for the scaling law being derived off of $16\times 16$ patch experiments, while this final budget is under $8\times 8$ patches, which requires roughly $4\times$ as much compute.} $\compute$ and the coefficients from \tabref{tab:opt-model-table} to \eqref{eqn:chinchilla}.

While we train with under half the compute, our scaling laws in Sec.~\ref{subsec:scalinglaws} predict equal performance with three times less compute. Thus, as predicted by our scaling laws and validated empirically by the results in \ref{tab:two-view-result} both SVSM models outperform decoder-only LVSM. Notably, SVSM-420M, the model trained in accordance with our scaling laws performs the best. For completeness, we also show reported results from prior work on this benchmark in \tabref{tab:two_view_STOA}. 

Furthermore, we also benchmark the \textit{rendering speed}, which is calculated with $V_T = 1$ to simulate real-time online rendering with respect to a stream of input poses. Additional details can be found in \suppref{supp:rendering}. As seen in \tabref{tab:two-view-result}, SVSM generally renders much faster than decoder-only LVSM, and  is eclipsed only once $V_C$ increases to $8$. 



\section{Scaling Laws for Multiview \boldmath\texorpdfstring{$\left(V_C{>}2\right)$}{(Vc>2)} NVS}
\label{sec:scaling-mv}

\paragraph{Analysis Setup.}
We continue to experiment in the multiview paradigm ($V_C{>}2$), which necessitates the reconciliation of scene information across many views to produce quality renderings. Specifically, we focus on $V_C = 4$ regime. For training and evaluation, we choose DL3DV~\citep{ling2024dl3dv}, a real-world dataset with wider baselines and more complex camera trajectories and subject matter, making it more suitable for multiview experiments. We follow the official test-train split and use $V_T{=}4$, and scene batch size of $64$ for all experiments to save resources. 
 All other settings follow those described in Sec.~\ref{sec:scaling-two-view} and \suppref{supp:exp-details}.
\begin{table*}[t]
  \centering
  \begin{threeparttable}
    \scriptsize
    \setlength{\tabcolsep}{3pt} 

    \begin{subtable}{\linewidth}
      \begin{tabular*}{\linewidth}{@{\extracolsep{\fill}} l *{9}{c}}
        \toprule
        
        & \multicolumn{3}{c}{\textbf{Scale Parameters}}
        & \multicolumn{3}{c}{\textbf{Reconstruction Quality}}
        & \multicolumn{3}{c}{\textbf{Rendering FPS (↑)}} \\
        
        \cmidrule(lr){2-4} \cmidrule(lr){5-7} \cmidrule(lr){8-10}
        
        \textbf{Model} 
        & \textbf{Model Size} & \textbf{Train Iters} & \textbf{Train FLOPs (↓)}  
        & \textbf{PSNR} (↑) & \textbf{SSIM} (↑) & \textbf{LPIPS} (↓)
        & $V_C$=4  & $V_C$=8  & $V_C$=16\\
        
        \midrule
        
        
        LVSM Decoder-only + PRoPE~\citep{jin2024lvsm,li2025cameras}
        & 171M & 100k & 43 eflops 
        & 26.19 & 0.830 & 0.145
        & 104.7 & 52.6 & 23.8 \\
        
        SVSM Enc-Dec (\textbf{ours}, $\approx$ Iter-matched)
        & 711M & 100k & 32 eflops
        & \underline{26.29} & \underline{0.835} & \underline{0.141} 
        & 280.4 & 261.2 & 230.4 \\

        SVSM Enc-Dec (\textbf{ours}, Pareto-optimal)
        & 400M & 233k & 44 eflops
        & \textbf{26.87} & \textbf{0.853} & \textbf{0.129}
        & \textbf{411.1} & \textbf{381.1} & \textbf{333} \\
        
        \bottomrule
      \end{tabular*}
    \end{subtable}

    \caption{
        \textbf{Multiview (}$V_C{>}2$\textbf{) NVS Results of the Largest Models.} Our compute-matched model achieves significantly better rendering quality ($+0.68\text{ PSNR}$, $-0.016\text{ LPIPS}$), while maintaining nearly four times the rendering speed at inference-time.
    }
    \label{tab:multiview-result} 
  \end{threeparttable}
\end{table*}
\begin{figure*}[t]
    \centering
    \includegraphics[width=\textwidth]{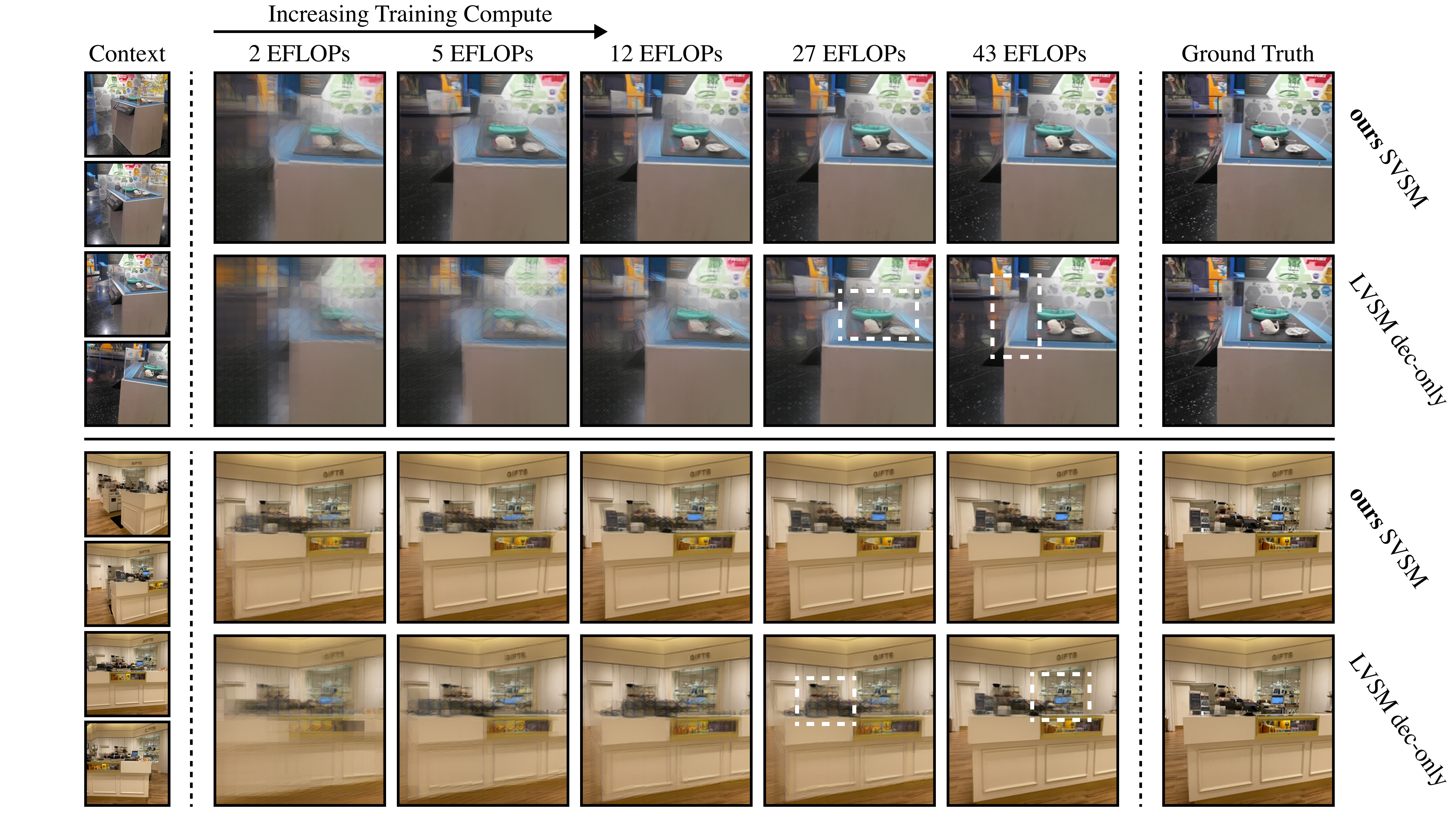}
    \caption{
        \textbf{Qualitative Scaling Behavior}, $V_C = 4$. The performance of both LVSM and our method steadily increases with compute (left to right). Compared vertically, for a given compute budget SVSM renderings are consistently less blurry.
    }
    \label{fig:qualitative-multiview}
\end{figure*}

\paragraph{Scaling Law Does Not Hold.}
Unfortunately, we find that naively extending our SVSM architecture to the multiview scenario does not result in a similar scaling trend. As can be seen from Figure~\ref{fig:mv_scale}a, the Pareto frontier of our unidirectional model saturates much quicker than bidirectional LVSM as we increase the train compute. 


\vspace{-1\baselineskip}
\paragraph{Relative Camera Attention Re-establishes Scaling Law.}
We hypothesize that this is not a fundamental problem of encoder-decoder paradigm, but a problem caused by the way our model utilizes the pose information. Specifically, we find that adding a form of relative camera attention~\citep{miyato2023gta,kong2024eschernet,li2025cameras} resolves this issue.

\looseness=-1
Let $\pose_i$ be the camera pose of the view that the $i$-th token belongs to. Relative camera attention models leverage attention mechanisms that only depend on the relative camera poses $\pose_{ij}{=}\pose_i^{-1}\pose_j$. 
This is typically achieved by 1)~mapping the query, key, and value vectors to an arbitrary global reference frame, 2)~performing attention there, and then 3)~mapping back the results to each token's own reference frame. This mechanism embeds the pose information directly into the attention layers, ensuring that it isn't lost after the initial embedding. This potentially explains the efficacy of the method for our model, which may lose the pose information through the bottleneck otherwise.
%



For our model, we adopt the recently proposed PRoPE~\citep{li2025cameras} embedding as the relative camera attention mechanism. We illustrate SVSM's architecture with the incorporation of PRoPE embeddings in \figref{fig:prope}. After adding PRoPE embeddings to both LVSM and SVSM, we retrain all models. Results are shown in \figref{fig:mv_scale}. The equivalent scaling is re-established, and SVSM again maintains a tighter Pareto-frontier. Qualitative scaling results for both models are shown in \figref{fig:qualitative-multiview}.
Note that while both models benefit from PRoPE embeddings, the advantage is far more pronounced in our encoder-decoder SVSM. 
\vspace{-1\baselineskip}

\paragraph{Final Models.}

Equipped with PRoPE embeddings, we again train larger models to compare directly against the 24-layer LVSM Decoder-only model, also equipped with PRoPE. As we did in the stereo case, we train a naive forward pass-matched model along with a Pareto-optimal model from a $N(\compute{})$ fit to the data. The performance of both models are listed in \tabref{tab:multiview-result} and their test loss curves are plotted in \figref{fig:mv_scale}. Again, the version of SVSM which follows the scaling laws outperforms both models, with significant $0.7$ PSNR and $-0.016$ LPIPS gaps. Beyond the superior reconstruction quality, the efficiency of SVSM becomes clear in the multi-view case, with a rendering FPS that is $4\times$ that of the decoder-only model, increasing to $14\times$ when extrapolated to larger context view counts.


\section{Scaling Laws for Fixed Latent Design}
\begin{figure}
    \centering
    \includegraphics[width=\linewidth]{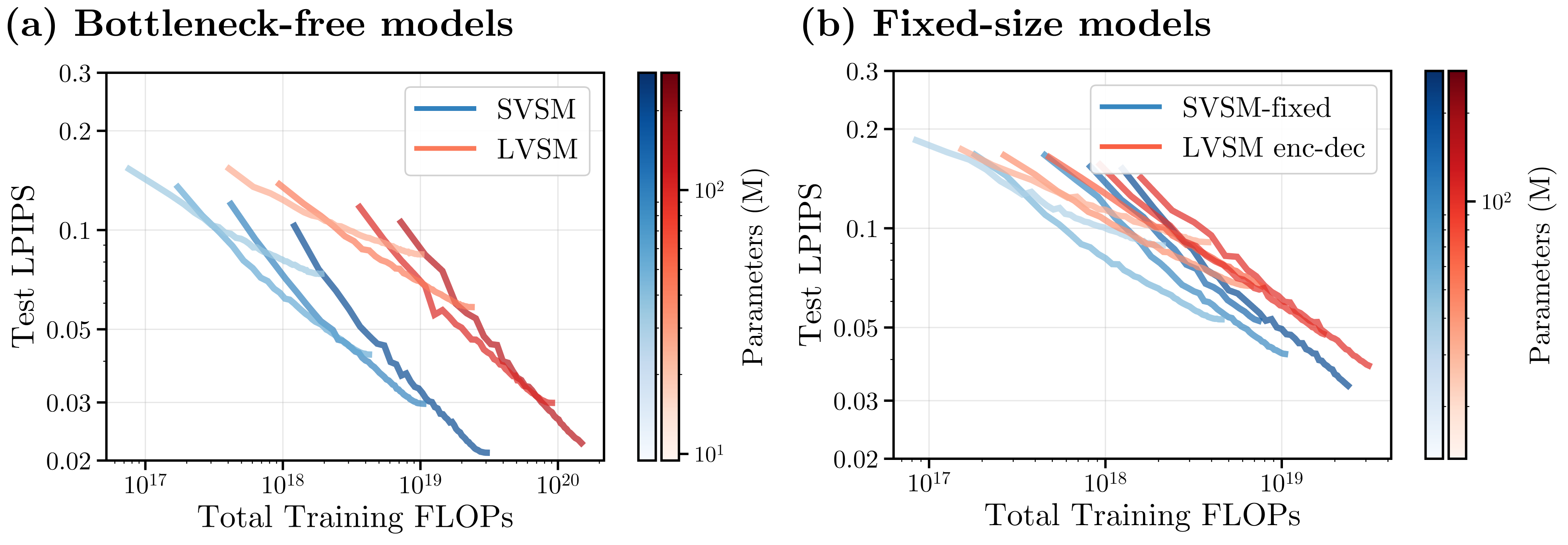}
    \caption{\textbf{Fixed-size Latent Scaling Experiments.} Conducted on Objaverse~\cite{objaverse}. \textbf{(a)} For $V_C{=}8$, SVSM and LVSM decoder-only scale equally while SVSM's frontier is shifted by $5\times$ on the compute axis. \textbf{(b)} When a fixed latent bottleneck is used, SVSM-fixed and LVSM encoder-decoder scale equally, but significantly worse than the unbottlenecked designs. SVSM again maintains a superior pareto frontier.}
    \label{fig:obj_scale}
\end{figure}

\paragraph{Analysis Setup.} Lastly, we check both the scaling laws and the design space of view synthesis models with a fixed-size scene representation. This design has favorable rendering speeds for large context lengths, though it does not have favorable training compute as the encoder is still quadratic in the context length. For training and evaluation, we use Objaverse~\citep{objaverse}, with 8 context views (where the benefit of having a fixed latent starts to appear at inference time). We compare two designs: LVSM Encoder-Decoder and SVSM-fixed, which follows the same design of a unidirectional decoder but instead decodes off of a fixed latent. Both models utilize PRoPE with identity pose on the scene representation. We use $V_T = 8$ and scene batch size $64$ for an effective batch size of $512$ for all experiments. All other settings follow those described in Sec.~\ref{sec:scaling-two-view} and \suppref{supp:exp-details}.

\paragraph{SVSM-Fixed Matches LVSM Scaling, but Both Scale Poorly.} As shown in Fig.~\ref{fig:obj_scale}b, SVSM with a fixed latent and the LVSM encoder–decoder exhibit similar scaling behavior. However, SVSM-fixed consistently requires less compute to achieve the same performance, maintaining a Pareto advantage. This indicates that our unidirectional decoder remains more compute-efficient even when a fixed latent bottleneck is imposed, which is desirable when amortized rendering is required. Nevertheless, comparing to Fig.~\ref{fig:obj_scale}a makes clear that both fixed-latent designs scale substantially worse than their bottleneck-free counterparts.
\section{Conclusion}
\label{sec:conclusion}
In this work, we established a rigorous compute-normalized benchmark for transformer view synthesis models.
 Our empirical studies reveal the importance of the concept of \emph{effective batch size} ---the product of the number of scenes in a batch with the number of per-scene rendering target views --- which redefines the notion of batch size for NVS training.
Based on this insight, we propose the Scalable View Synthesis Model (SVSM), which features a unidirectional encoder-decoder architecture for favorable scaling with effective batch. We demonstrate that SVSM is dramatically more compute-efficient than the current SOTA architecture, LVSM, and consistently achieves the same performance with $2-3\times$ less training compute.
We further demonstrate that relative camera pose embeddings in multi-view attention is the key to realizing favorable scaling behavior with increasing numbers of context views. 
Lastly, we show that even with a fixed-size latent representation, our unidirectional decoder is still more compute-efficient than the LVSM encoder–decoder architecture; however, both approaches scale substantially worse than the designs without a latent bottleneck.
In sum, our findings establish a new framework for evaluating the performance and effectiveness of transformer view synthesis models.  

\paragraph{Acknowledgements.} This work was supported by the National Science Foundation under Grant No. 2211259, by the Singapore DSTA under DST00OECI20300823 (New Representations for Vision, 3D Self-Supervised Learning for Label-Efficient Vision), by the Intelligence Advanced Research Projects Activity (IARPA) via Department of Interior/ Interior Business Center (DOI/IBC) under 140D0423C0075, by the Amazon Science Hub, by the MIT-Google Program for Computing Innovation, and by a 2025 MIT Office of Research Computing and Data Seed Grant.

{
    \small
    \bibliographystyle{ieeenat_fullname}
    \bibliography{main}
}

\clearpage
\maketitlesupplementary

\section{FLOPs for View Synthesis Transformers}
\label{supp:flop-calcs}

In this section, we explain how we calculated FLOP consumption for all models. Additionally, we show that in our regime, the MLP cost is most significant. In a vision transformer, there are three contributions to the FLOPs:
\begin{itemize}
    \item Initial patchifying + tokenization layers (negligible)
    \item Transformer blocks: attention.
    \item Transformer blocks: MLP and projections.
\end{itemize}
The first is negligible as it is a single linear layer at the start. Letting $n$ be the number of tokens and $d$ the transformer dimension, for each self-attention transformer block, the following are the FLOPs consumed by the attention, projection, and MLP layers:
\begin{align}
    \computeMLP, \computeProj &\propto n d^2 \\
    \computeAttn &\propto n^2 d
\end{align}
Thus, the total FLOPs consumed for a transformer block are of the form
\begin{equation}
    \compute{} = A n^2 d + B nd^2,
\end{equation}
where $A$ is $4$ while $B$ is $16$ in the self-attention case, so even at the smallest model size list (\suppref{supp:size-list}), $\frac{Bnd^2}{An^2d} \approx 4 \cdot \frac{384}{512} \approx 3$, meaning in most cases our MLP / linear projection FLOPs take up a large majority. For SVSM's cross-attention decoder, the formula becomes a little more complicated as there are separate $n_{\text{ctxt}}$ and $n_{\text{target}}$, but these remain on the same order of magnitude as $n$, so the same is true. 

For the decoder-only model, the number of tokens $n$ is given by
\begin{equation}
    n = (V_C + 1) \cdot \frac{H}{p} \cdot \frac{W}{p},
\end{equation}
where $p$ is the patch size and $H$ and $W$ are the width and height. For the SVSM encoder-decoder, the number of active tokens vary as 
\begin{equation}
    n_{\text{enc}} = V_C \cdot \frac{H}{p} \frac Wp, n_{\text{dec}} = V_T \cdot \frac Hp \cdot \frac Wp,
\end{equation}
So, for both models we can write down the MLP and attention complexities as:
\begin{align}
    \computeMLP^{(\text{LVSM})} &\propto V_T n \propto V_T (V_C + 1) \\
    \computeMLP^{(\text{SVSM})} &\propto n_{\text{enc}} + n_{\text{dec}} \propto V_C + V_T \\
    \computeAttn^{(\text{LVSM})} &\propto V_T n^2 \propto V_T (V_C + 1)^2 \\
    \computeAttn^{(\text{SVSM})} &\propto n_{\text{enc}}^2 + n_{\text{enc}}n_{\text{dec}} \propto V_C^2 + V_CV_T
\end{align}
For fixed $V_C$, we see that multiplying $V_T$ by an amount $k$ scales compute by exactly $k\times$ in LVSM, while it only scales compute by $\frac{kV_T + V_C}{V_T + V_C}$ in the SVSM case, which is where its advantage lies.

\begin{table}[t]
\centering
\begin{threeparttable}
\footnotesize            
\setlength{\tabcolsep}{4pt}  

\begin{tabular}{lcccccc}   
\toprule

& \multicolumn{3}{c}{\textbf{Encoder}}
& \multicolumn{3}{c}{\textbf{Decoder}} \\
\cmidrule(lr){2-4} \cmidrule(lr){5-7}

\textbf{Params}
& \textbf{dim} & \textbf{dim\_head} & \textbf{n\_layers}
& \textbf{dim} & \textbf{dim\_head} & \textbf{n\_layers} \\
\midrule


15M & 384 & 64 & 3 & 384 & 64 & 3 \\
27M & 384 & 64 & 6 & 384 & 64 & 6 \\
35M & 384 & 64 & 8 & 384 & 64 & 8 \\
62M & 512 & 64 & 8 & 384 & 64 & 8 \\
79M & 512 & 64 & 8 & 512 & 64 & 12 \\
145M & 640 & 64 & 10 & 640 & 64 & 14 \\
226M & 768 & 64 & 10 & 768 & 64 & 16 \\
316M & 768 & 64 & 12 & 768 & 64 & 24 \\
\hline
\textbf{420M} & 768 & 64 & 16 & 768 & 64 & 32 \\
740M & 1024 & 64 & 16 & 1024 & 64 & 32 \\

\bottomrule
\end{tabular}

\caption{\textbf{RealEstate10K $\boldsymbol{V_C}\mathbf{{=}2}$, SVSM Encoder-Decoder.} SVSM Encoder-Decoder model settings used to sweep scaling laws for the stereo ($V_C{=} 2$) novel view synthesis setting. \textbf{Bolded} is the row for the compute-matched, Pareto-optimal SVSM model compared against the 24-layer LVSM Decoder-only model.}
\label{tab:svsm-re10k-details}
\end{threeparttable}
\end{table}
\begin{table}[t]
\centering
\begin{threeparttable}

\footnotesize            
\setlength{\tabcolsep}{4pt}  

\begin{tabular}{lcccccc}   
\toprule

& \multicolumn{3}{c}{\textbf{Encoder}}
& \multicolumn{3}{c}{\textbf{Decoder}} \\
\cmidrule(lr){2-4} \cmidrule(lr){5-7}

\textbf{Params}
& \textbf{dim} & \textbf{dim\_head} & \textbf{n\_layers}
& \textbf{dim} & \textbf{dim\_head} & \textbf{n\_layers} \\
\midrule

  8M & - & - & - & 384 & 64 & 3 \\
13M & - & - & - & 384 & 64 & 6 \\
22M & - & - & - & 512 & 64 & 6 \\
28M & - & - & - & 512 & 64 & 8 \\
53M & - & - & - & 640 & 64 & 10 \\
90M & - & - & - & 768 & 64 & 12 \\
118M & - & - & - & 768 & 64 & 16 \\
175M & - & - & - & 768 & 64 & 24 \\
275M & - & - & - & 896 & 64 & 28 \\

\bottomrule
\end{tabular}

\caption{\textbf{RealEstate10K $\boldsymbol{V_C}\mathbf{{=}2}$, LVSM Decoder-only.} LVSM Decoder-only model settings used to sweep scaling laws for the stereo ($V_C{=}2$) novel view synthesis setting.}
\label{tab:dec-only-re10k-details}
\vspace{-\baselineskip}
\end{threeparttable}
\end{table}

\section{Further Experimental Details}
\label{supp:exp-details}

\subsection{Training Details}

All models in this study are multi-view ViTs, following the design of LVSM~\citep{jin2024lvsm}. The main deviation is no layer index-dependent initialization. All layers are initialized with the same standard deviation. Instead, for stability, we apply a $1/\sqrt{L}$ multiplier on the residuals of all layers where $L$ is the depth of the transformer. Empirically, we find this to maintain stable training on the same learning rate across multiple transformer depths. 

All models are trained with the AdamW optimizer, with a peak learning rate of $4\text{e-}4$, $\beta_1 = 0.9$, $\beta_2 = 0.95$, and weight decay of $0.05$ on all parameters except LayerNorm weights, all following LVSM. We additionally warmup the learning rate for $3000$ steps in all models. 

All models are trained on $256\times 256$ resolution for both DL3DV and RealEstate10K. All models are trained with the same reconstruction metric used in~\citep{jin2024lvsm}. In particular, the loss is
\begin{equation}
    \mathcal L = \text{MSE}(I_T, \tilde I_T) + \lambda \cdot \text{Perceptual}(I_T,\tilde I_T),
\end{equation}
where we choose $\lambda = 0.5$ as our perceptual loss weight. 

We also have a few experiment specific details. For $V_C{=}2$, RealEstate10K, we sample our context and target views from a video index range from 25-192. For $V_C{>}2$, DL3DV, we sample our context and target views from a video index range of 16-24 as the baselines between consecutive frames in DL3DV are much wider. For effective batch tests under $V_C{=}4$ on DL3DV, we train each model for 100k iterations, while under $V_C{=}2$ on RealEstate10K, we train for less iterations (50k) to save compute resources. The trend is still clear even with the limited iterations. 

\begin{table}[t]
\centering
\begin{threeparttable}

\footnotesize            
\setlength{\tabcolsep}{4pt}  

\begin{tabular}{lcccccc}   
\toprule

& \multicolumn{3}{c}{\textbf{Encoder}}
& \multicolumn{3}{c}{\textbf{Decoder}} \\
\cmidrule(lr){2-4} \cmidrule(lr){5-7}

\textbf{Params}
& \textbf{dim} & \textbf{dim\_head} & \textbf{n\_layers}
& \textbf{dim} & \textbf{dim\_head} & \textbf{n\_layers} \\
\midrule

15M & 384 & 64 & 3 & 384 & 64 & 3 \\
32M & 384 & 64 & 6 & 384 & 64 & 8 \\
85M & 512 & 64 & 10 & 512 & 64 & 12 \\
168M & 640 & 64 & 12 & 640 & 64 & 16 \\
280M & 768 & 64 & 12 & 768 & 64 & 20 \\
711M & 1024 & 64 & 24 & 1024 & 64 & 24 \\
\hline
\textbf{400M} & 768 & 64 & 24 & 768 & 64 & 24 \\

\bottomrule
\end{tabular}

\caption{\textbf{DL3DV $\boldsymbol{V_C}\mathbf{{=}4}$, SVSM Encoder-Decoder.} SVSM Encoder-Decoder model settings used to sweep scaling laws for the multi-view ($V_C > 2$) novel view synthesis setting. \textbf{Bolded} is the row for the compute-matched, Pareto-optimal SVSM model compared against the 24-layer LVSM Decoder-only model.}
\label{tab:svsm-dl3dv-details}
\end{threeparttable}
\end{table}

\begin{table}[t]
\centering
\begin{threeparttable}

\footnotesize            
\setlength{\tabcolsep}{4pt}  

\begin{tabular}{lcccccc}   
\toprule

& \multicolumn{3}{c}{\textbf{Encoder}}
& \multicolumn{3}{c}{\textbf{Decoder}} \\
\cmidrule(lr){2-4} \cmidrule(lr){5-7}

\textbf{Params}
& \textbf{dim} & \textbf{dim\_head} & \textbf{n\_layers}
& \textbf{dim} & \textbf{dim\_head} & \textbf{n\_layers} \\
\midrule

8M & - & - & - & 384 & 64 & 3 \\
22M & - & - & - & 512 & 64 & 6 \\
43M & - & - & - & 640 & 64 & 10 \\
90M & - & - & - & 768 & 64 & 12 \\
175M & - & - & - & 768 & 64 & 24 \\
383M & - & - & - & 1024 & 64 & 30 \\

\bottomrule
\end{tabular}

\caption{\textbf{DL3DV $\boldsymbol{V_C}\mathbf{{=}4}$, LVSM Decoder-only.} LVSM Decoder-only model settings used to sweep scaling laws for the multi-view ($V_C > 2$) novel view synthesis setting.}
\vspace{-0.5cm}
\label{tab:dec-only-dl3dv-details}
\end{threeparttable}
\end{table}

\begin{figure}[!t]
    \centering
    \includegraphics[width=\linewidth]{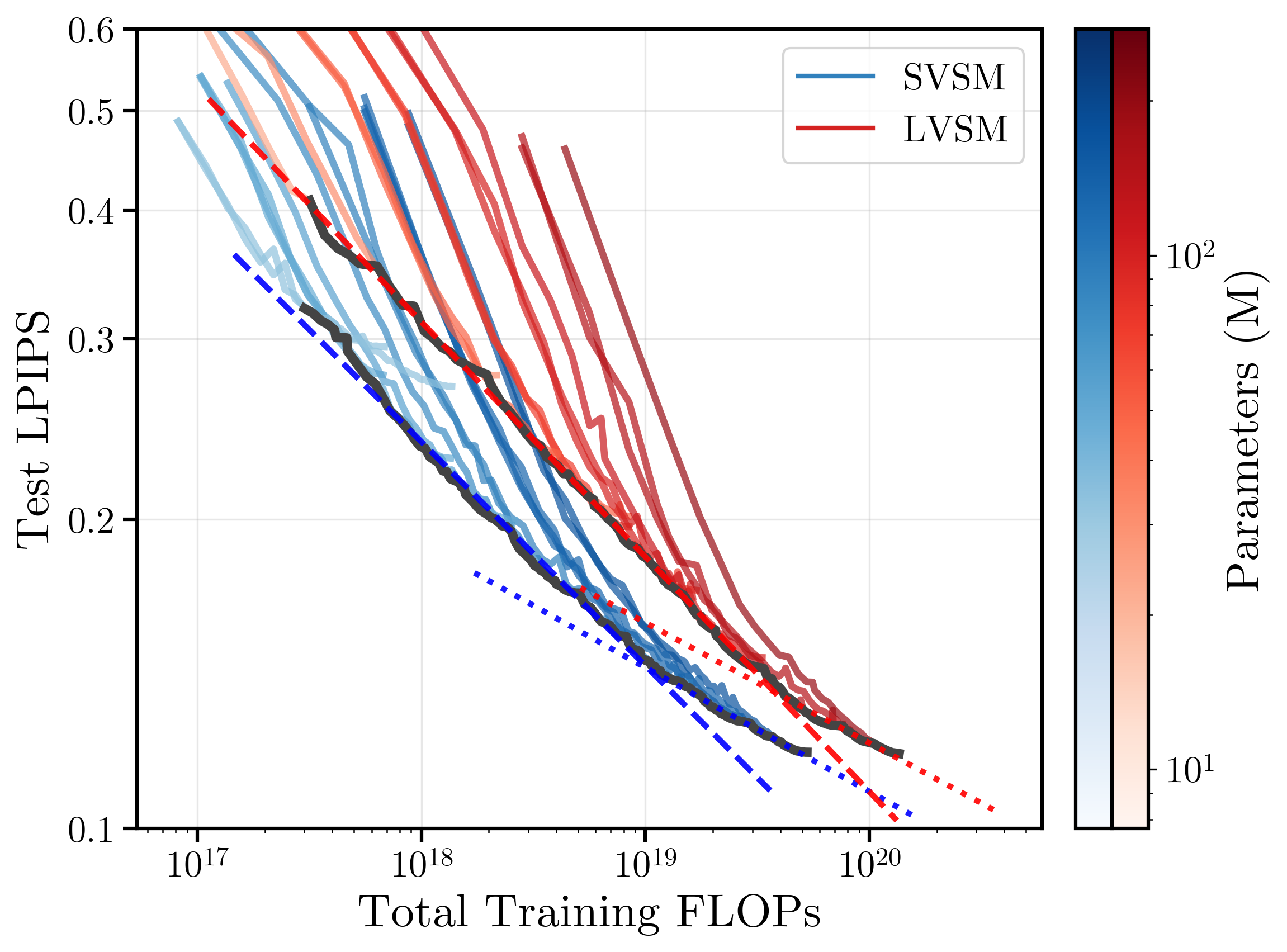}
    \vspace{-0.5cm}
    \caption{\textbf{Linear Power Scaling Laws.} We fit scaling laws onto sections of the Pareto-frontiers of the model families. We see that both models have approximately the same slope in each of their corresponding sections, indicating equal scaling.}
    \vspace{-0.5cm}
    \label{fig:fits}
\end{figure}

\subsection{Rendering Speed Benchmarking}
\label{supp:rendering}

We benchmark all rendering on a single A6000 GPU. We found that due to some hardware configurations of our setup, when benchmarking with batch size $1$, all models would cap out at 30 fps, even when the width of each layer was increased or decreased, suggesting that there was some non-FLOP based bottleneck. To circumvent this, all models were tested with batch size $64$, which allowed the rendering FPS to properly reflect the forward pass FLOPs for each of the models. We still used $V_T = 1$ for all models, and report \begin{equation}
    \text{FPS} = \frac{B \cdot V_T}{t_{\text{iter}}} = \frac{B}{t_{\text{iter}}},
\end{equation}
where $t_{\text{iter}}$ is the iteration time for one batch. If higher $V_T$ was used (offline rendering), the rendering speed difference between our model and LVSM would be even larger.

\subsection{Model Size List for Scaling Sweeps}
\label{supp:size-list}

We trained a wide range of models across both families and both problem settings. Their sizes and hyperparameters are listed in tables \ref{tab:svsm-re10k-details}, \ref{tab:dec-only-re10k-details}, \ref{tab:svsm-dl3dv-details}, and \ref{tab:dec-only-dl3dv-details}. For the decoder-only model there is not much room for flexibility in terms of model hyperparameters -- we simply scale dimension up along with the layer count. For SVSM encoder-decoder we can flexibly allocate different amounts of compute to the encoder and the decoder. For low context view cases, we allocate more to the decoder as there are less complex relations amongst the context views and for higher context views we allocate similar amounts to the decoder as to the encoder. Empirically, we roughly found this to have better performance-per-compute, but we did not thoroughly study this split, so we leave this to future work. 

\section{Linear Fits on the Loss vs. Compute Frontiers}
\label{supp:fits}

Though the trend is not perfectly linear, we fit lines onto sections of  $P(\compute{}) \propto \compute{}^c$ in Fig.~\ref{fig:fits}, which show equal scaling between SVSM and LVSM. The actual coefficients are reported in \tabref{tab:lin-fits}, which have almost identical power law coefficients across the two model families.

\begin{table}[h]
\footnotesize
\centering
\begin{tabular}{lcc}
\toprule
\textbf{Model} & $c$ for $P > 0.14$ & $c$ for $P \le 0.14$ \\
\midrule
LVSM & -0.23 & -0.12 \\
SVSM & -0.22 & -0.12 \\
\bottomrule
\end{tabular}
\caption{\textbf{LPIPS vs. compute scaling coefficients.} As regressed from  \figref{fig:fits}, we find power law coefficients for LPIPS $P(\compute{}) \propto \compute{}^c$, and we report $c$ in this table for the first half, which is determined by test LPIPS loss greater than $0.14$ and the second half which is test LPIPS loss less than $0.14$.
\vspace{-1\baselineskip}
}
\label{tab:lin-fits}
\end{table}

\section{PRoPE Ablations}

We conduct a series of ablations on PRoPE~\cite{li2025cameras} in the multi-view setting in an attempt to elucidate the mechanism through which it enables scaling. One potential source of success is that the PRoPE SVSM design allows for the decoder to see the clean context poses, while vanilla SVSM does not. To test if this might be the cause for the success, we concatenated context plucker rays to the scene representation from the encoder. However, this has negligible impact (Tab.~\ref{tab:prope-abl-1}). Additionally, replacing PRoPE with GTA~\cite{miyato2023gta} showed negligible difference, indicating epipolar geometry is not crucial for viewcount scalability. Hence, we presume that the relative embeddings are the key, i.e., canonicalizing features to the target frame. Lastly, having PRoPE on just the decoder performs nearly as well as having PRoPE on both, indicating that the cross attention seems to benefit the most from the relative embedding inductive bias.

\begin{table}[h]
\footnotesize
\centering
\begin{tabular}{lccc}
\toprule
\textbf{Model} & \textbf{PSNR} (↑) & \textbf{SSIM} (↑) & \textbf{LPIPS} (↓)\\
\midrule
Vanilla SVSM & 23.50 & 0.727 & 0.254 \\
Vanilla w/ concat. pose & 23.47 & 0.726 & 0.254 \\
\midrule
PRoPE on Encoder & 23.61 & 0.733 & 0.249 \\
PRoPE on Decoder & 24.31 & 0.771 & 0.220\\
PRoPE on both & 24.62 & 0.782 & 0.210 \\
GTA on both & 24.63 & 0.782 & 0.207 \\
\bottomrule
\end{tabular}
\caption{\textbf{PRoPE ablations.} We vary where we apply PRoPE, try a different relative attention method (GTA), and also test pose information flow. GTA varies negligibly from PRoPE, indicating that epipolar geometry is not crucial, and the skip pose connection also has negligible impact, indicating that pose information flow is not responsible.}
\vspace{-1\baselineskip}
\label{tab:prope-abl-1}
\end{table}


\section{Compiled Scaling Results}

\begin{figure}[!t]
    \centering
    \includegraphics[width=\linewidth]{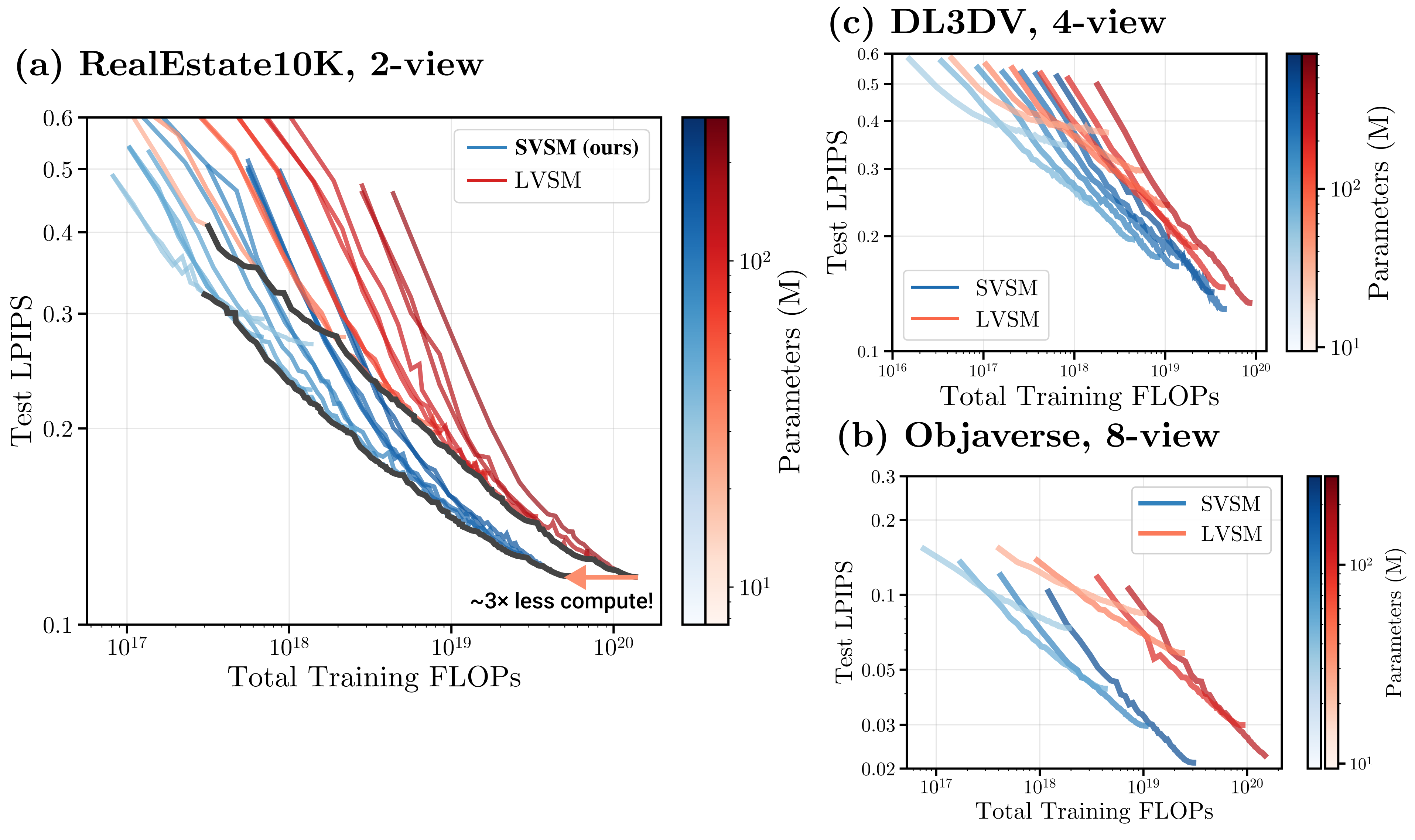}
    \vspace{-0.5cm}
    \caption{\textbf{All Scaling Laws.} We collect scaling laws across the three datasets we tested on, and in all cases, SVSM is compute-optimal.}
    \vspace{-0.5cm}
    \label{fig:collected_scaling}
\end{figure}

We provide a collection of scaling results across RealEstate10K, DL3DV, and Objaverse across 2, 4, and 8 context views respectively in Fig.~\ref{fig:collected_scaling}. In all cases, SVSM (in blue) maintains a pareto advantage over LVSM decoder-only.

\section{Further Qualitative Results}
\label{supp:qual}

\begin{figure*}[t]
    \centering
    \includegraphics[width=\textwidth]{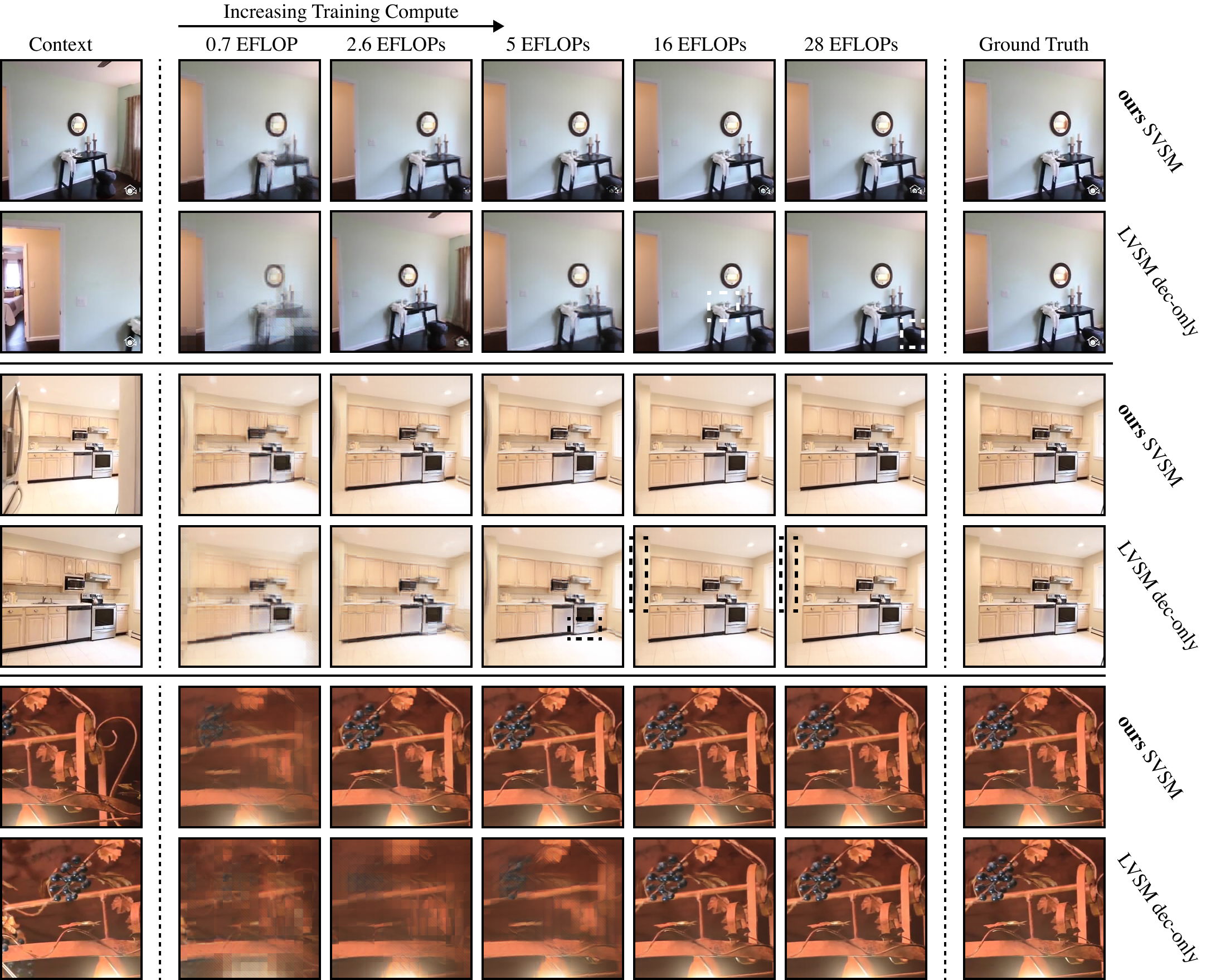}
    \caption{Qualitative results on RE10K ($V_C{=}2$) across scale.}
    \label{fig:supp-two-view-scale}
\end{figure*}

\begin{figure*}[t]
    \centering
    \includegraphics[width=\textwidth]{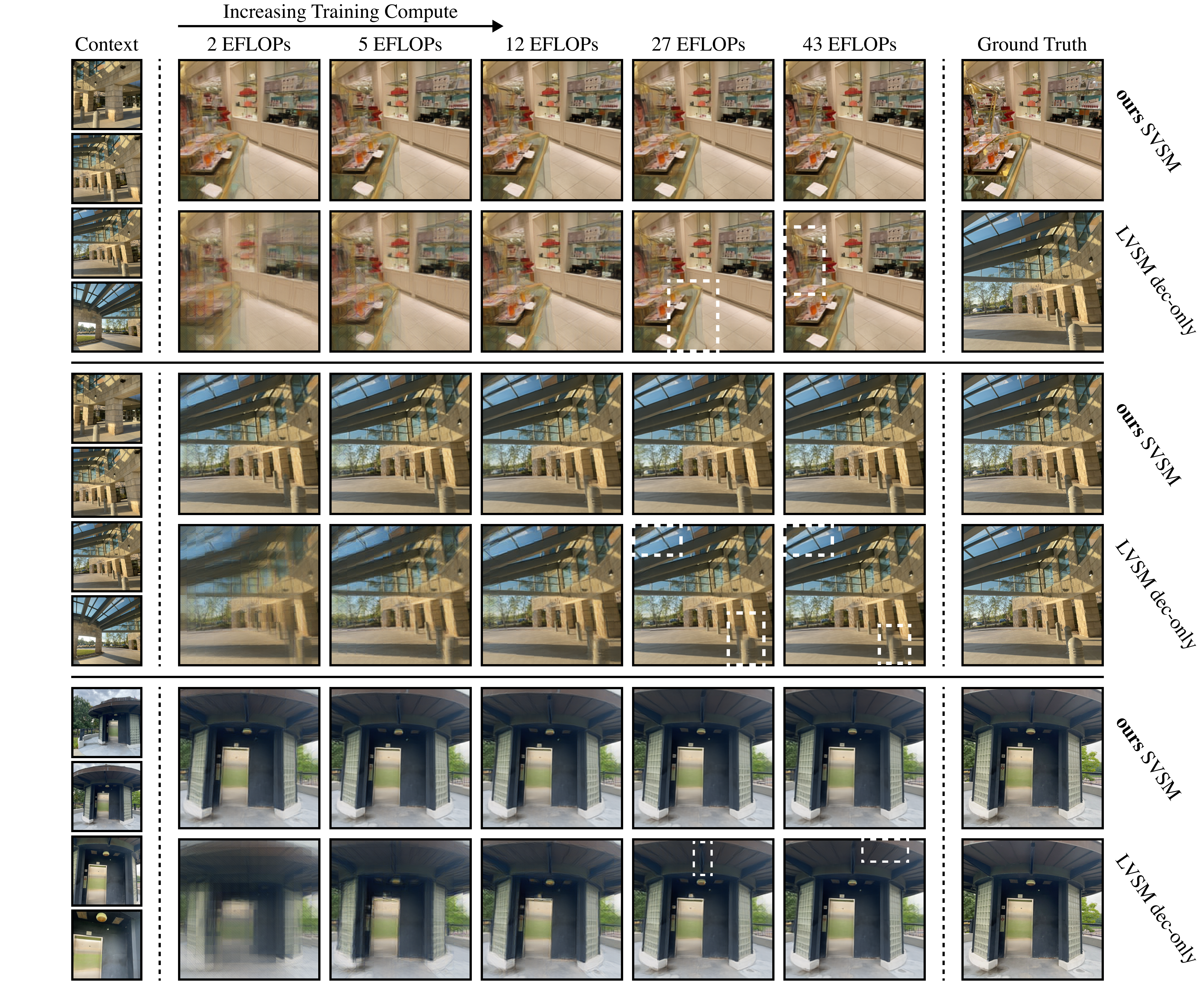}
    \caption{Qualitative results on DL3DV ($V_C{=}4$) across scale.}
    \label{fig:supp-four-view}
\end{figure*}

\begin{figure*}[t]
    \centering
    \includegraphics[width=\textwidth]{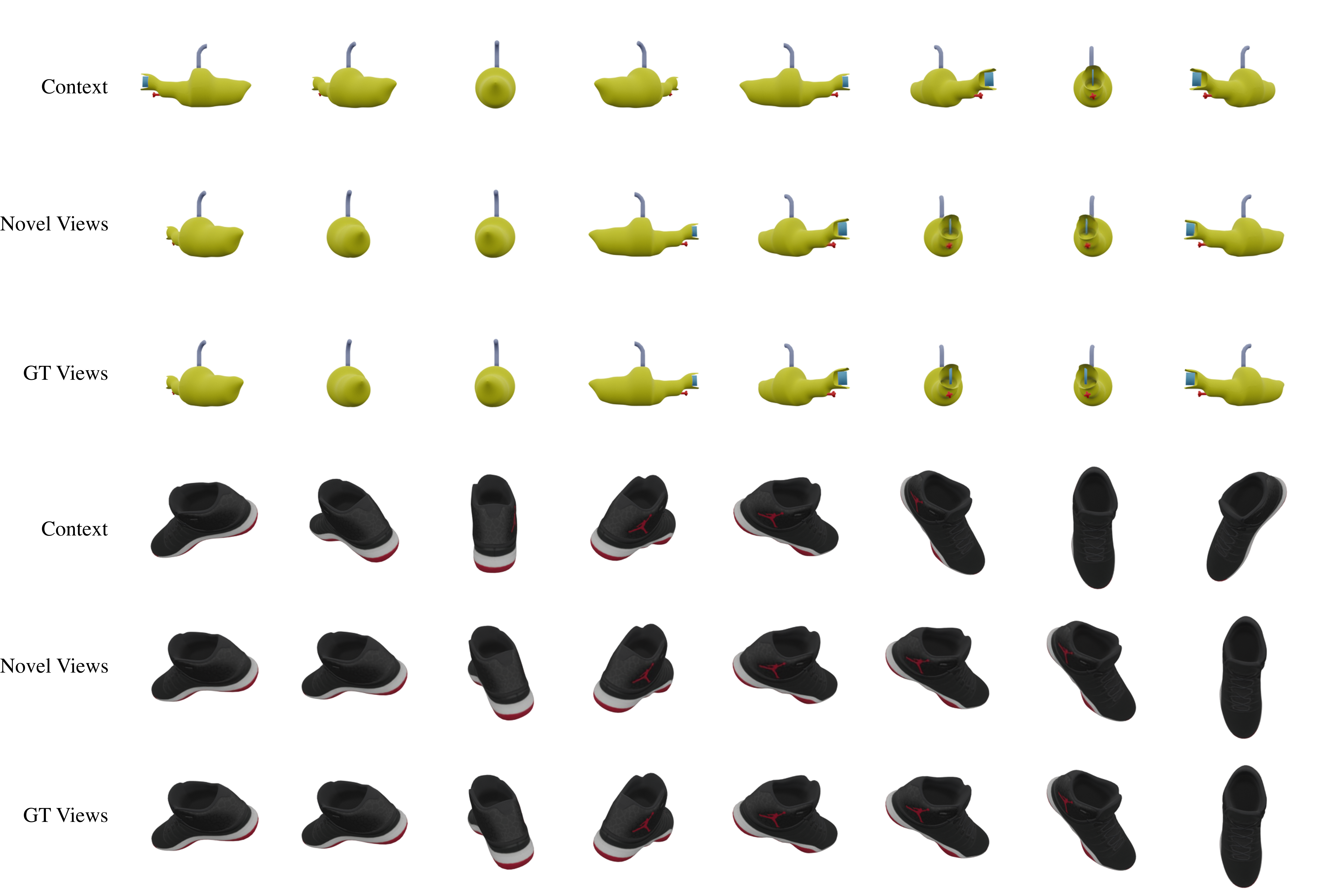}
    \caption{
        Multiview consistency results on Objaverse ($V_C{=}8$).
    }
    \label{fig:qualitative-multiview-objaverse}
\end{figure*}

Lastly, we provide more qualitative results across various training compute budgets of 2 context view evaluations on RealEstate10K in Fig.~\ref{fig:supp-two-view-scale} and 4 context view evaluations on DL3DV in Fig.~\ref{fig:supp-four-view}. Multiview consistency of outputs is shown in Fig.~\ref{fig:qualitative-multiview-objaverse} on Objaverse.



\clearpage

\end{document}